\documentclass[dvipsnames,table]{article}

\usepackage[final]{hunyuan}

\usepackage{amsmath}
\usepackage{amssymb}
\usepackage{mathtools}
\usepackage{amsthm}
\usepackage{multirow}
\usepackage{makecell}
\usepackage{subcaption}
\usepackage{wrapfig}
\usepackage{graphicx}

\usepackage{pifont}
\usepackage{fvextra}
\usepackage{xspace}

\usepackage[utf8]{inputenc} 
\usepackage[T1]{fontenc}    
\usepackage{url}            
\usepackage{booktabs}       
\usepackage{amsfonts}       
\usepackage{nicefrac}       
\usepackage{microtype}      
\usepackage{colortbl}
\usepackage{color, soul}
\usepackage[table]{xcolor}
\usepackage{booktabs}
\usepackage{textgreek}

\usepackage[labelfont=bf]{caption}
\usepackage{enumitem}
\usepackage{sectsty}
\usepackage{pgfplots}
\pgfplotsset{compat=1.17}
\usepackage{fancyhdr}
\usepackage{natbib} 
\let\cite\citep    

\renewcommand{\headrulewidth}{1pt}
\makeatletter
\def\headrule{{\if@fancyplain\let\headrulewidth\plainheadrulewidth\fi
\hrule\@height\headrulewidth\@width\textwidth \vskip-\headrulewidth}}
\makeatother

\definecolor{HYDarkBlue}{HTML}{2155EA} 
\definecolor{HYLightBlue}{HTML}{A8DFF6}
\definecolor{HYLightBlueLighter}{HTML}{C6ECFA}

\definecolor{syh}{RGB}{41, 91, 160}
\usepackage[colorlinks, linkcolor=HYDarkBlue, anchorcolor=HYDarkBlue, citecolor=HYDarkBlue, urlcolor=HYDarkBlue]{hyperref}

\usepackage{cleveref}
\usepackage{threeparttable}

\sectionfont{\color{HYDarkBlue}\fontfamily{zi4}\selectfont}
\subsectionfont{\color{HYDarkBlue}\fontfamily{zi4}\selectfont}
\subsubsectionfont{\color{HYDarkBlue}\fontfamily{zi4}\selectfont}

\newcommand{\method}{\textbf{AngelSpec}\xspace}

\newcommand{\dcut}{\textbf{D-cut}\xspace}
\newcommand{\dflaretwo}{\textbf{DFly}\xspace}

\title{\method: Towards Real-World High Performance Inference with Speculative Decoding}

\author{%
\textbf{\normalsize{Hong Liu, \quad Rui Cen,\quad Junhan Shi,\quad Guangshuo Qin, \quad Jiebin Zhang, \quad Tianyu Liu, \quad Runzhi Fan}}\\
\textbf{\normalsize{Guoliang Zhao, \quad Ruobing Xie, \quad Kai Zhang, \quad Song Liu,\quad Guanghua Yu\textsuperscript{*},\quad Jianchen Zhu}}\\[0.4em]
{\normalfont\small Tencent Inc.}\\
{\normalfont\small \textsuperscript{*}Corresponding author}
\vspace{0em}
}

\def\github{\raisebox{-1.5pt}{\includegraphics[height=1.05em]{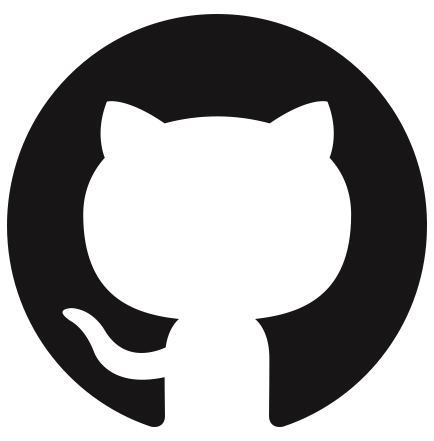}}}
\def\email{\raisebox{-1.5pt}{\includegraphics[height=1.05em]{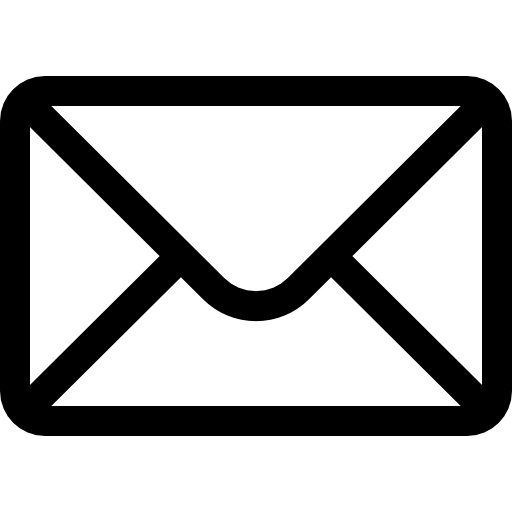}}}
\def\blog{\raisebox{-1.5pt}{\includegraphics[height=1.05em]{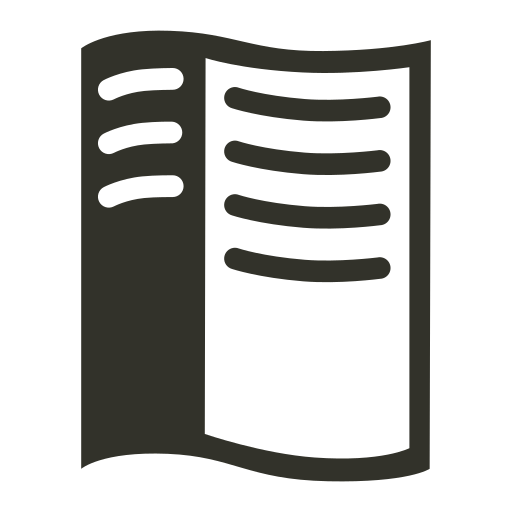}}}

\usepackage{geometry}
\usepackage{algorithm}
\usepackage{algpseudocode}
\usepackage{setspace}

\makeatletter
\renewcommand{\ALG@beginalgorithmic}{\small}
\makeatother



\begin{document}

\maketitle
\thispagestyle{fancy} 


\begin{abstract}
Speculative decoding can accelerate large language model inference without changing the target distribution, but no single drafting structure performs best across real-world workloads.
Autoregressive multi-token prediction (MTP) provides a lightweight and stable proposal mechanism, whereas block-parallel diffusion amortizes drafting latency over much longer candidate sequences.
Their advantages depend strongly on the output distribution, and \method addresses this workload heterogeneity from training, architecture, and inference perspectives.
At the training level, instead of optimizing one universal drafter on a uniform data mixture, we jointly specialize model structure and training data: an MTP drafter is trained on rich and diverse conversation-oriented data for high-entropy, open-ended chat, while a block-diffusion drafter is strengthened with code- and mathematics-focused data to exploit longer predictable continuations.
At the architecture level, we propose \dflaretwo, a block-diffusion framework that combines a hybrid target-conditioning backbone with a predecessor-conditioned autoregressive head, improving both target-feature utilization and intra-block dependency modeling while preserving high-throughput parallel generation.
At the inference level, we observe that acceptance lengths vary across domains, requests, and decoding steps, while the cost of verifying additional tokens changes with online load and deployment hardware.
\dflaretwo therefore treats verification as a shared batch-level resource: it reallocates compute toward high-confidence prefixes across requests and combines their expected utility with a profiled runtime cost model to adapt verification depth to the current serving condition. We present \method, a unified training framework for MTP and block-parallel speculative decoding, and conduct systematic experiments across the Hy3 model series.
On Hy3-A21B, \dflaretwo increases the average accepted length by approximately $30\%$, enabling the highest average throughput at every tested concurrency from 4 to 64, delivering a $1.98$--$2.40\times$ speedup over autoregressive decoding and $10.5$--$11.8\%$ higher throughput than DFlash. We release the \method framework to support training and extending these speculative-decoding methods.

\end{abstract}

\section{Introduction}
\label{sec:introduction}

Large language models (LLMs) generate one token per autoregressive forward pass, making decoding increasingly expensive as model scale and serving demand grow.
Speculative decoding reduces this cost without changing the target distribution: a lightweight drafter proposes multiple future tokens, and the target model verifies them together in one forward pass using rejection sampling~\citep{sps1,sps2,xia2024survey}.
The verified prefix and a target-generated bonus token are then committed, allowing one decoding round to advance by multiple tokens.
Its acceleration therefore depends jointly on the number of accepted draft tokens and the latency of the complete draft--verify round.

Most speculative-decoding studies search for one drafter that performs well on an averaged benchmark mixture.
However, real-world serving does not follow such a uniform distribution.
Online requests span open-ended conversation, code generation, mathematical reasoning, tool use, and agentic workflows, whose conditional entropy and continuation structure differ substantially.
Our measurements on the Hy3 model family reveal that these differences materially change which drafting paradigm is preferable.
In high-entropy conversation, many continuations are semantically valid, but the target may select any one of them.
Acceptance therefore decays quickly with proposal depth, limiting the value of generating and verifying a long block.
For this regime, autoregressive MTP drafting provides a favorable balance because it proposes a shorter candidate sequence and avoids over-speculating when only a few draft positions are likely to be useful~\citep{mtp}.
By contrast, code and mathematical reasoning exhibit a more constrained structure.
Programming syntax, repeated identifiers, formal expressions, and step-by-step derivations constrain future tokens more strongly and often create longer predictable spans.
These workloads can benefit substantially from block-parallel drafting, which predicts an entire candidate block in one backbone pass and amortizes drafting latency over many positions.
DFlash demonstrates the potential of this direction by using a diffusion-style drafter conditioned on target hidden states~\citep{dflash}.
However, using the same general-purpose data and the same drafting structure for all domains leaves considerable performance untapped.
We instead specialize both \emph{model structure} and \emph{training distribution}: MTP is trained with rich and diverse conversation-oriented data, whereas our block-diffusion model is strengthened with code- and mathematics-focused samples.

For block-diffusion drafting, we propose \dflaretwo with two architectural improvements on DFlash~\citep{dflash}.
First, its hybrid target-conditioning backbone combines a globally transformed target context with layer-specific target views, allowing different draft layers to use target features suited to their depth rather than sharing one fixed representation~\citep{zhang2026dflare}.
Second, its predecessor-conditioned autoregressive head corrects each parallel prediction using the token selected at the preceding draft position~\citep{dspark,huang2026domino,rheinboldt2026treeflashparallelarapproximationfaster}.
This correction turns position-wise marginal predictions into prefix-conditioned distributions, improving intra-block dependency modeling while retaining a single-pass parallel backbone.
Together with domain-strengthened training data, these changes improve both target-feature utilization and suffix coherence for block-diffusion drafting.

At inference time, acceptance length varies across domains, requests, and decoding steps, while the cost of verifying additional tokens changes with online load, model architecture, parallelism strategy, and deployment hardware.
A fixed verification depth may therefore underuse predictable drafts under light load or waste limited batch capacity on low-value suffixes under high concurrency~\citep{illusion,echo}.
\dcut addresses this serving-side uncertainty by treating target verification as a shared batch-level resource: the system reallocates compute toward high-confidence prefixes across concurrent requests and combines their expected acceptance utility with a profiled runtime cost model.
This allows verification depth to adapt to the current serving condition, retaining longer drafts when target compute is available and pruning aggressively when verification becomes expensive.

To support these designs, we present \method, an open-source training framework for MTP and block-parallel speculative decoding.
Training the two drafter paradigms on large MoE targets places demands that existing recipes do not cover: both consume target hidden states at scale, MTP additionally requires an on-policy TTT rollout inside the training step and long-context training, and \dflaretwo requires per-layer target conditioning with pluggable autoregressive heads.
\method meets these with a disaggregated design in which inference engines generate target hidden states and stream them directly to training workers, together with document-aware sequence packing and an evaluation server that measures genuine acceptance behavior during training.
Using \method, we conduct systematic experiments across the Hy3 model series, covering chat, code, mathematics, and serving conditions with different concurrency levels.
The results demonstrate that workload specialization, the architectural improvements in \dflaretwo, and runtime-adaptive verification provide complementary gains.
We release the \method framework to support training, evaluating, and extending these speculative-decoding methods.

Our main contributions are:
\begin{itemize}[leftmargin=1.5em,itemsep=2pt]
    \item We identify workload heterogeneity as a first-class design consideration for real-world speculative decoding and train complementary drafters: a conversation-oriented MTP model for high-entropy chat and a domain-strengthened \dflaretwo for code and mathematics.
    \item We propose \dflaretwo, a block-diffusion framework that combines a hybrid target-conditioning backbone with a predecessor-conditioned autoregressive head, improving target-feature utilization and intra-block dependency modeling while retaining a parallel backbone.
    \item We integrate \dcut to allocate target-verification compute across requests according to prefix confidence and runtime cost, adapting verification depth to diverse acceptance behavior, online load, and deployment hardware. On Hy3-A21B, \dflaretwo achieves the highest average throughput at every tested concurrency from 4 to 64, delivering a $1.98$--$2.40\times$ speedup over autoregressive decoding and $10.5$--$11.8\%$ higher throughput than DFlash.
    \item We develop and release \method, an open-source training framework with unified support for MTP and block-parallel speculative decoding---covering disaggregated hidden-state generation, TTT rollout, long-context training, and online acceptance evaluation---and systematically evaluate these methods across the Hy3 model family and diverse serving conditions.
\end{itemize}
\section{Multi-Token Prediction}
\label{sec:mtp}

Multi-token prediction (MTP) augments next-token training with auxiliary predictions of future tokens and has been adopted by a growing number of production model families~\citep{mtp,deepseekai2026deepseekv4,zeng2026glm}.
Unlike a separately trained small language model, an MTP module is attached to the target backbone and directly reuses its hidden states, vocabulary embedding, and language-model head.
It therefore provides a low-overhead starting point for self-speculation without introducing a separate drafter model.

The original Hy3 model is trained with a single MTP layer and without recurrent self-conditioned unrolling.
During inference, this block can be reused recurrently to propose several tokens, but the original training objective does not prepare it for a long self-generated chain.
The mismatch becomes important after the first draft position.
The original MTP block is trained for one teacher-forced application with ground-truth tokens and clean hidden states, whereas multi-step inference repeatedly applies the same block to its own predicted tokens, recurrent hidden states, and growing draft KV cache.
Errors therefore accumulate with depth, causing the second and third draft positions to have substantially lower acceptance rates than the first.

We address this train--inference mismatch by applying TTT to the original MTP block in a shared-parameter, multi-depth training scheme.
We retain $D$ logical prediction depths indexed by $k\in\{0,\ldots,D-1\}$, but all depths reuse the same physical MTP block.
During training, this block is autoregressively unrolled for $D$ steps, with each prediction fed into the next invocation.
Although the parameters are shared, depth $k$ is supervised by its own shifted future-token target.
Training-Time Test (TTT) therefore exposes the shared MTP block to the self-generated trajectories encountered during inference, while memory-efficient distillation aligns every depth with the corresponding frozen target distribution.
This shared MTP block is the short-horizon drafter in \method, complementing the longer block-parallel drafter in \Cref{sec:dflare}.

\subsection{Shared-Parameter Multi-Depth MTP Drafter}
\label{sec:mtp-overview}

\begin{figure}[t]
    \centering
    \includegraphics[width=\textwidth]{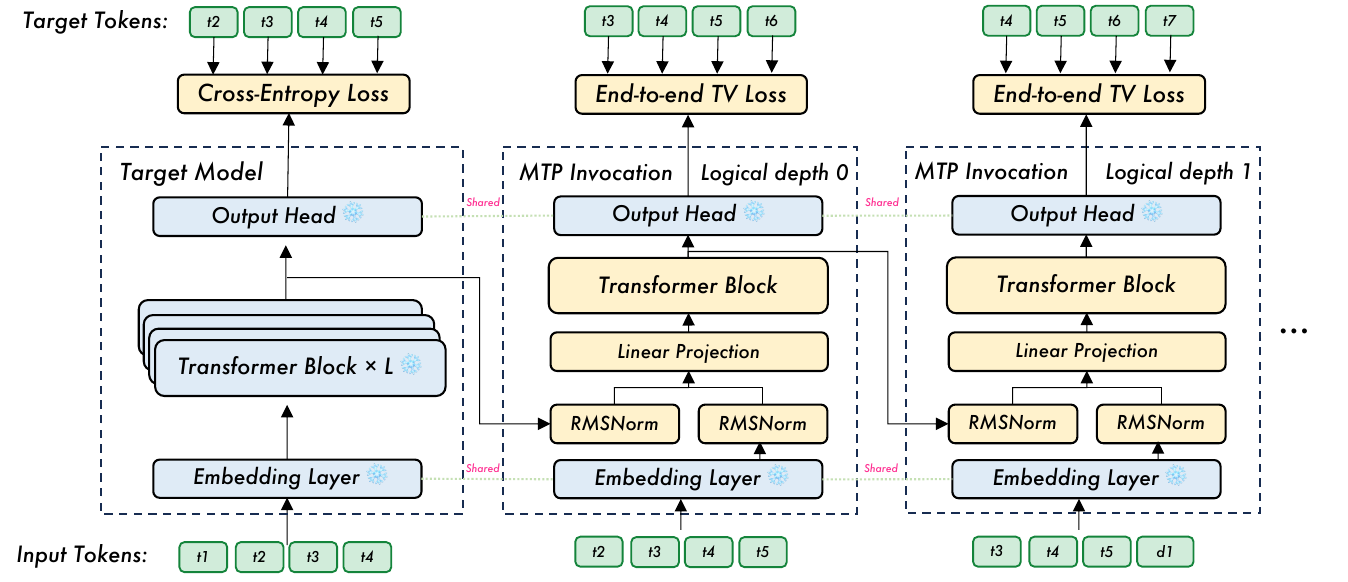}
    \caption{Shared-parameter multi-depth MTP training with Training-Time Test. }
    \label{fig:mtp-arch}
\end{figure}

\Cref{fig:mtp-arch} shows the target model and the first two logical prediction depths.
These depths are separate points in the autoregressive rollout, but they share one MTP module and one set of parameters.
For a drafting round beginning at online position $t$, let $h^{(-1)}$ denote the target hidden state entering the round and let $\widetilde{x}_0$ be the token aligned with the first MTP input.
Here, $t$ indexes the online drafting position, while the superscript $k$ indexes the zero-based logical prediction depth.
For readability, the online-position subscript $t$ is omitted from the variables in the following equations.
At prediction depth $k\in\{0,\ldots,D-1\}$, the shared MTP block receives the hidden state from its preceding invocation and the latest available input-token embedding:
\begin{equation}
    u^{(k)}
    =
    W_{\mathrm{proj}}
    \left[
        \operatorname{RMSNorm}\!\left(h^{(k-1)}\right);
        \operatorname{RMSNorm}\!\left(E\!\left(\widetilde{x}_{k}\right)\right)
    \right],
    \label{eq:mtp-fusion}
\end{equation}
where $E$ is the target token embedding, $\widetilde{x}_0$ is the token aligned with the first MTP invocation, and, for $k\geq 1$, $\widetilde{x}_{k}$ is either the shifted ground-truth token or the draft token selected at depth $k-1$.
The fused feature passes through one Transformer layer,
\begin{equation}
    h^{(k)}
    =
    F_{\theta}
    \left(
        u^{(k)},\mathcal{K}^{(<k)}
    \right),
    \qquad
    q_k
    =
    \operatorname{softmax}
    \left(
        W_{\mathrm{lm}}h^{(k)}
    \right),
    \label{eq:mtp-depth}
\end{equation}
where $\mathcal{K}^{(<k)}$ denotes the draft-side KV states accumulated before depth $k$ and $W_{\mathrm{lm}}$ is the target language-model head.
The selected token $\widehat{x}_k$ from $q_k$ becomes $\widetilde{x}_{k+1}$ during self-conditioned rollout.
The $D$ predictions form a short linear draft block that the target verifies in one forward pass.

All prediction depths share the same MTP parameters $\theta$ and projection $W_{\mathrm{proj}}$; the repeated MTP boxes in \Cref{fig:mtp-arch} denote recurrent invocations rather than independent modules.
The token embedding and output head are also shared with the target.
The supervision is nevertheless depth-specific: the target logits are shifted by one additional position at every depth, so $q_k$ is matched with a different teacher distribution.
During training, the target backbone and target language-model head are frozen, and the MTP inputs from the backbone are detached.
The drafter can therefore improve its proposal distribution without changing the target distribution that speculative verification must preserve.

\subsection{Training-Time Test}
\label{sec:mtp-ttt}

Standard teacher forcing does not reproduce the state distribution encountered by a multi-step drafter.
For a valid training position $i$, depth $k=0$ uses the same aligned input token in training and inference.
At prediction depth $k\geq 1$, however, teacher-forced training supplies the ground-truth token $x_{i+k+1}$, whereas inference must use the sampled or greedy prediction $\widehat{x}_{i+k+1}$ produced at depth $k-1$.
The hidden state and draft KV cache consequently drift away from the clean training trajectory after the first error.
This exposure bias is absent at depth $k=0$ but compounds over the remainder of the draft chain.

Following the Training-Time Test principle of EAGLE-3~\citep{eagle3}, we autoregressively unroll the same MTP module during training.
For $k\in\{0,\ldots,D-2\}$, the module at depth $k{+}1$ receives the $\arg\max$ prediction produced by its depth-$k$ invocation instead of the ground-truth token.
At the same time, a growing draft KV cache and a causal-prefix plus diagonal-draft attention pattern reproduce the dependency structure used at inference.
The same parameters consequently receive training signals from several autoregressive depths, while the teacher target advances by one future position at each depth.
Later depths are therefore trained to recover from realistic upstream errors rather than assuming a perfect prefix.
\paragraph{Long-context training.}
TTT increases training memory because every prediction depth maintains rollout state and compares its student distribution with a differently shifted target distribution.
At long sequence lengths, the full-vocabulary teacher and student logits dominate this overhead.
To scale TTT to long contexts, we shard the sequence dimension and keep rollout states, attention computation, and logits distribution matching distributed throughout training.
For packed examples, all depth-dependent shifts and validity masks respect document boundaries, preventing supervision from leaking between unrelated sequences.
This design bounds per-device memory without replicating full sequence-level tensors on any single worker.
\subsection{Acceptance-Aligned Distillation}
\label{sec:mtp-loss}

Let $i$ index a token position in a training sequence, and let $\mathcal{I}_k$ denote the set of positions with valid supervision at zero-based prediction depth $k$.
The frozen target head provides a distinct teacher distribution $p_{i,k}$ at every valid position and prediction depth, while the shared MTP block produces the student distribution $q_{i,k}$.
In the implementation, the teacher logits are cumulatively shifted by one token at each depth.
Equivalently, at sequence position $i$ and zero-based depth $k$, the MTP prediction is aligned with token $x_{i+k+2}$.
Thus, the parameters are shared across depths, but the prediction target changes with depth.
The forward-KL baseline is
\begin{equation}
    \mathcal{L}_{\mathrm{KL}}
    =
    \sum_{k=0}^{D-1}
    w_k
    \frac{1}{|\mathcal{I}_k|}
    \sum_{i\in\mathcal{I}_k}
    D_{\mathrm{KL}}
    \left(
        p_{i,k}\,\|\,q_{i,k}
    \right),
    \label{eq:mtp-kl}
\end{equation}
where $w_k$ is a normalized per-depth weight and the loss at each depth is averaged over valid token positions.
To bound the memory of fp32 softmax intermediates, we compute the distribution-matching terms on the teacher's top-$K$ support.

KL is stable, but it remains a proxy for speculative acceptance.
To directly measure the mismatch between the target distribution $p$ and the MTP distribution $q$, we use the total variation (TV) distance:
\begin{equation}
    \operatorname{TV}(p,q)
    =
    \frac{1}{2}
    \sum_{v\in\mathcal{V}}
    \left|p(v)-q(v)\right|,
    \label{eq:mtp-tv}
\end{equation}
where $\mathcal{V}$ is the token vocabulary.
TV measures the probability mass that must be moved to transform one distribution into the other: it is zero when $p=q$ and approaches one as their overlap vanishes.
Unlike KL divergence, TV is directly connected to speculative-sampling acceptance.
Under rejection-sampling verification, the expected single-position acceptance probability is the distributional overlap
\begin{equation}
    \alpha(p,q)
    =
    \sum_{v\in\mathcal{V}}\min\!\left(p(v),q(v)\right)
    =
    1-\operatorname{TV}(p,q).
    \label{eq:mtp-acceptance-tv}
\end{equation}
The LK Losses framework proposes two objectives that retain this direct connection to acceptance~\citep{lk-losses}.
The first adaptively blends KL and TV,
\begin{equation}
    \mathcal{L}_{\mathrm{LK}}^{\lambda}(p,q)
    =
    \lambda D_{\mathrm{KL}}(p\|q)
    +
    (1-\lambda)\operatorname{TV}(p,q),
    \qquad
    \lambda
    =
    \exp\!\left(-\eta\,\operatorname{sg}[\alpha]\right),
    \label{eq:mtp-lk}
\end{equation}
where $\operatorname{sg}$ denotes stop-gradient.
When the drafter is poorly aligned, KL supplies a smooth, high-magnitude training signal.
As overlap improves, the objective shifts weight toward TV, which directly targets rejection-sampling acceptance.
The second is the negative log-overlap objective
\begin{equation}
    \mathcal{L}_{\mathrm{LK}}^{\alpha}(p,q)
    =
    -\log\!\left(\alpha(p,q)+\epsilon\right),
    \label{eq:mtp-lk-alpha}
\end{equation}
Since $\alpha(p,q)=1-\operatorname{TV}(p,q)$ and the target distribution $p$ is fixed, the chain rule gives
\begin{equation}
    \nabla_\theta \mathcal{L}_{\mathrm{LK}}^{\alpha}
    =
    -\frac{1}{\alpha+\epsilon}\nabla_\theta\alpha
    =
    \frac{1}{\alpha+\epsilon}
    \nabla_\theta\operatorname{TV}(p,q_\theta),
    \label{eq:mtp-lk-alpha-gradient}
\end{equation}
at differentiable points, with the same relation holding for compatible subgradients elsewhere.
Thus, the negative log-overlap objective preserves the per-token TV gradient direction while assigning it a larger weight when the distributional overlap is low.
When $\epsilon\ll\alpha$, the scaling factor is approximately $1/\alpha$.
We use this negative log-overlap LK Loss as the cold-start objective in our final TTT configuration.

After cold start, training switches to the end-to-end TV objective\cite{li2026breaking} based on the expected accepted draft length, where $\alpha_{i,k}=\alpha(p_{i,k},q_{i,k})$:
\begin{equation}
    \mathcal{L}_{\mathrm{e2e}}
    =
    1-
    \frac{1}{|\mathcal{I}|D}
    \sum_{i\in\mathcal{I}}
    \sum_{m=0}^{D-1}
    \prod_{k=0}^{m}\alpha_{i,k},
    \label{eq:mtp-e2e}
\end{equation}
where $\mathcal{I}$ contains the training positions with valid supervision across all $D$ depths.
The distribution-matching terms are computed using $K=10000$.
This multiplicative form gives shallow positions more influence because a rejection at an early position invalidates every later draft token.
\Cref{eq:mtp-acceptance-tv} is the distributional training motivation, while the results in \Cref{sec:mtp-results} use the observed acceptance statistic under their respective greedy or stochastic decoding settings.
\paragraph{Loss ablation.}
\Cref{tab:mtp-loss-ablation} reports a controlled loss ablation on a small-scale MoE target model with draft depth $D=4$.
KL consistently improves over hard-label CE, and LK further improves the mean accepted length averaged across benchmarks.
The e2e TV configuration with LK Loss cold start, which first optimizes LK Loss and then switches to e2e TV Loss, achieves the best result on all the benchmarks.
In contrast, directly optimizing per-token TV from the initial checkpoint without LK Loss cold start is unstable because the TV gradient is weak when the student distribution is far from the target.
Based on this comparison, we adopt e2e TV Loss with LK Loss cold start as the final loss configuration.

\begin{table*}[t]
    \centering
    \setlength{\tabcolsep}{3.5pt}
    \renewcommand{\arraystretch}{1.12}
    \def\hlg{\cellcolor{HYLightBlue!55}}
    \caption{\textbf{MTP loss ablation.} Each entry is the expected accepted length, including one target-generated bonus token, for a small-scale MoE target model with draft depth $D=4$; Avg. is the mean over the seven displayed benchmarks. The architecture and training data are fixed. The TV Loss configuration directly optimizes the per-token TV objective, whereas the LK-cold-start configuration first optimizes LK Loss and then switches to e2e TV Loss.}
    \label{tab:mtp-loss-ablation}
    \resizebox{\textwidth}{!}{%
    \begin{tabular}{l*{8}{>{\normalsize}c}}
        \toprule
        \multirow{2}{*}{\textbf{Training objective}} &
        \multicolumn{2}{c}{\textbf{Math}} &
        \multicolumn{3}{c}{\textbf{Code}} &
        \multicolumn{2}{c}{\textbf{Chat}} &
        \multirow{2}{*}{\textbf{Avg.}} \\
        \cmidrule(lr){2-3}\cmidrule(lr){4-6}\cmidrule(lr){7-8}
        & {\small\textbf{GSM8K}} & {\small\textbf{Math500}}
        & {\small\textbf{HumanEval}} & {\small\textbf{MBPP}}
        & {\small\textbf{LiveCodeBench}} & {\small\textbf{AlpacaEval}}
        & {\small\textbf{MT-Bench}} & \\
        \midrule
        CE Loss
            & $4.09$ & $4.23$ & $4.00$ & $3.92$ & $3.69$ &
              $3.08$ & $3.60$ & $3.80$ \\
        KL Loss
            & $4.20$ & $4.32$ & $4.12$ & $4.03$ & $3.82$ &
              $3.24$ & $3.72$ & $3.92$ \\
        LK Loss
            & $4.23$ & $4.34$ & $4.13$ & $4.06$ & $3.86$ &
              $3.27$ & $3.75$ & $3.95$ \\
        TV Loss
            & $3.59$ & $3.53$ & $2.95$ & $2.91$ & $2.59$ &
              $2.72$ & $2.94$ & $3.03$ \\
        e2e TV Loss (with LK cold start)
            & \hlg $\mathbf{4.25}$ & \hlg $\mathbf{4.36}$ & \hlg $\mathbf{4.15}$ &
              \hlg $\mathbf{4.08}$ & \hlg $\mathbf{3.88}$ &
              \hlg $\mathbf{3.27}$ & \hlg $\mathbf{3.77}$ &
              \hlg $\mathbf{3.96}$ \\
        \bottomrule
    \end{tabular}%
    }
\end{table*}

\subsection{Training Data Strategy}
\label{sec:mtp-data}

MTP post-training is highly sensitive to the training distribution.
The drafter is not learning an independent language model; it is learning to reproduce the target model's future-token behavior under the contexts that will appear at serving time.
A narrow corpus may improve acceptance on matched benchmarks while leaving large gaps on other workloads.
We therefore organize the training data around three principles.

\paragraph{Diverse task coverage.}
The training mixture should contain a broad range of sequence structures and output distributions, including open-ended conversation, knowledge-intensive question answering, mathematical reasoning, code generation, instruction following, and agentic interaction.
These domains stress different aspects of the shared MTP module.
Code and mathematics provide strongly constrained continuations and clear supervision for deeper prediction depths, whereas conversation and knowledge tasks expose the drafter to higher-entropy target distributions.
Training on all of them prevents the shared parameters from specializing only to the easiest low-entropy continuations and improves robustness across downstream evaluations.

\paragraph{Long-context and code-agent trajectories.}
Short examples are insufficient for learning the state distribution encountered after a long prompt or a long generation history.
We therefore include long-context data, particularly code-agent trajectories that contain repository context, iterative analysis, tool interaction, code modification, and multi-step feedback.
Such samples expose the MTP module to long-range dependencies, repeated identifiers, delayed constraints, and changes in generation mode within one sequence.
This coverage is important for maintaining acceptance at long context lengths; increasing the configured sequence length alone cannot compensate for the absence of representative long-context trajectories.

\paragraph{Target-model rollout.}
We obtain stronger supervision by generating responses from the frozen target model and training MTP on these target-generated responses, rather than directly using the original reference responses.
The original response may have been written by a human or generated by another model and can follow a token distribution that differs from the deployed target.
In contrast, target rollout produces the exact token choices, hidden-state trajectories, and local uncertainty patterns that MTP must approximate during speculative decoding.
It therefore reduces the teacher--data mismatch and gives better acceptance than training on the unmodified source responses.
In practice, we first assemble a diverse prompt mixture, generate completions with the target model under the intended serving format, and then apply TTT to these target-aligned trajectories.

\subsection{Experimental Results}
\label{sec:mtp-results}

\paragraph{Metrics and setup.}
For draft position $i$, $p_i$ is its cumulative acceptance rate under target verification at the selected decoding temperature.
We report $\mathrm{Avg}=(p_1+p_2+p_3)/3$ and the mean accepted length (MAL).
With $D=3$ and one target-generated bonus token, $\operatorname{MAL}=1+\sum_{i=1}^{3}p_i$.

The main comparison uses vLLM at draft depth $D=3$ under greedy decoding ($T=0$) and stochastic decoding ($T=0.9$) on seven benchmarks spanning mathematics (GSM8K and Math500), code (HumanEval, MBPP, and LiveCodeBench), and open-ended chat scenarios (MT-Bench and AlpacaEval).

\paragraph{Main results.}
\Cref{tab:mtp-main} compares the base MTP checkpoint with the checkpoint obtained after TTT on the diverse target-model rollout mixture.
At $T=0$, average acceptance increases from $52.8\%$ to $66.4\%$, a gain of $13.6$ percentage points, and MAL rises from $2.58$ to $2.99$.
At $T=0.9$, the same model improves average acceptance from $51.3\%$ to $63.3\%$ and MAL from $2.54$ to $2.90$.
The improvement at both temperatures is concentrated in the speculative suffix.
At $T=0$, the aggregate rates change from $0.799/0.518/0.266$ to $0.814/0.653/0.524$ for $p_1/p_2/p_3$; at $T=0.9$, they change from $0.789/0.499/0.253$ to $0.792/0.619/0.486$.
Thus, the first-position rate is nearly preserved, while the second and third positions improve substantially.
For example, $p_3$ increases from $0.290$ to $0.706$ on GSM8K and from $0.387$ to $0.757$ on HumanEval.
This is the desired behavior for short-horizon drafting: the post-training recipe preserves the already-strong first proposal and converts previously low-value deeper positions into accepted progress.

\begin{table*}[t]
    \centering
    \small
    \setlength{\tabcolsep}{3.5pt}
    \renewcommand{\arraystretch}{1.2}
    \setlength{\aboverulesep}{0pt}
    \setlength{\belowrulesep}{0pt}
    \def\hlg{\cellcolor{HYLightBlue!55}}
    \caption{\textbf{Hy3 MTP acceptance at $T=0$ and $T=0.9$.} Base and TTT+Rollout rows report Avg/MAL, where Avg is the mean of the three cumulative position-wise acceptance rates. $\Delta$Avg rows report the improvement in Avg in percentage points.}
    \label{tab:mtp-main}
    \resizebox{\textwidth}{!}{%
    \begin{tabular}{llcccccccc}
        \toprule
        \multirow{2}{*}{\textbf{Temperature}} &
        \multirow{2}{*}{\textbf{Model}} &
        \multicolumn{2}{c}{\textbf{Math}} &
        \multicolumn{3}{c}{\textbf{Code}} &
        \multicolumn{2}{c}{\textbf{Chat}} &
        \multirow{2}{*}{\textbf{Mean}} \\
        \cmidrule(lr){3-4}\cmidrule(lr){5-7}\cmidrule(lr){8-9}
        & & \textbf{GSM8K} & \textbf{Math500}
        & \textbf{HumanEval} & \textbf{MBPP} & \textbf{\shortstack{LiveCodeBench}}
        & \textbf{AlpacaEval} & \textbf{MT-Bench} & \\
        \midrule
        \multirow{3}{*}{$T=0$}
        & Base
        & $56.8\%/2.70$ & $58.2\%/2.75$ & $68.1\%/3.04$
        & $58.4\%/2.75$ & $54.0\%/2.62$ & $35.1\%/2.05$
        & $38.8\%/2.16$ & $52.8\%/2.58$ \\
        & \textbf{TTT+Rollout}
        & \hlg $\mathbf{80.6\%/3.42}$ & \hlg $\mathbf{71.7\%/3.15}$ & \hlg $\mathbf{85.1\%/3.55}$
        & \hlg $\mathbf{71.3\%/3.14}$ & \hlg $\mathbf{63.1\%/2.89}$ & \hlg $\mathbf{44.4\%/2.33}$
        & \hlg $\mathbf{48.5\%/2.45}$ & \hlg $\mathbf{66.4\%/2.99}$ \\
        & $\Delta$Avg
        & $+23.8$ & $+13.5$ & $+17.0$ & $+12.9$ & $+9.1$
        & $+9.3$ & $+9.7$ & $+13.6$ \\
        \midrule
        \multirow{3}{*}{$T=0.9$}
        & Base
        & $54.8\%/2.65$ & $55.5\%/2.67$ & $66.3\%/2.99$
        & $57.2\%/2.71$ & $53.6\%/2.61$ & $34.2\%/2.03$
        & $37.8\%/2.13$ & $51.3\%/2.54$ \\
        & \textbf{TTT+Rollout}
        & \hlg $\mathbf{76.9\%/3.31}$ & \hlg $\mathbf{68.2\%/3.05}$ & \hlg $\mathbf{83.0\%/3.49}$
        & \hlg $\mathbf{66.5\%/2.99}$ & \hlg $\mathbf{59.8\%/2.79}$ & \hlg $\mathbf{42.5\%/2.28}$
        & \hlg $\mathbf{45.9\%/2.38}$ & \hlg $\mathbf{63.3\%/2.90}$ \\
        & $\Delta$Avg
        & $+22.1$ & $+12.7$ & $+16.7$ & $+9.3$ & $+6.2$
        & $+8.3$ & $+8.1$ & $+11.9$ \\
        \bottomrule
    \end{tabular}%
    }
\end{table*}

\section{DFly: Scalable and AR-Aware Parallel Drafting}
\label{sec:dflare}

Parallel drafters generate all $\gamma$ candidate tokens in one backbone forward pass, making drafting latency nearly independent of block length and enabling substantially longer proposals than autoregressive drafters.
DFlash is a representative design: it projects hidden states from multiple target layers into a shared context, injects that context into every draft layer through the KV sequence, and predicts all masked block positions with bidirectional attention~\citep{dflash}.
This parallelism provides strong throughput, but the shared target context limits layer specialization and the predicted positions lack explicit autoregressive dependencies.

\dflaretwo addresses these limitations through three coordinated improvements.
First, a hybrid target-conditioning backbone combines DFlash's expressive shared projection with DFlare's layer-specific target views~\citep{zhang2026dflare}.
Second, a lightweight autoregressive head introduces predecessor information after the parallel backbone.
Third, an acceptance-aware objective aligns distribution matching with the contribution of each block position to the expected accepted length.

\subsection{Hybrid Target-Conditioning Backbone}
\label{sec:dflarev2-backbone}

\dflaretwo builds on two complementary target-conditioning structures.
DFlash concatenates hidden states from multiple target layers and transforms them with a fully connected layer, producing one shared context feature at position $t$:
\begin{equation}
    c_t
    =
    W_{\mathrm{fc}}
    [h_t^{(1)};\ldots;h_t^{(T)}]
    +b_{\mathrm{fc}},
    \label{eq:dflarev2-dflash-feature}
\end{equation}
where $h_t^{(j)}\in\mathbb{R}^{d}$ is the hidden state from the $j$-th selected target layer.
The resulting context sequence
$C=[c_1;\ldots;c_L]$ contains a strong nonlinear transformation of multi-level target information, but the same $C$ is supplied to every draft layer~\citep{dflash}.

DFlare takes a different approach.
For a draft model with $D$ layers, it learns
$W_{\mathrm{fuse}}\in\mathbb{R}^{D\times T}$ and constructs a layer-specific weighted sum:
\begin{equation}
    \alpha^{(i)}
    =
    \operatorname{softmax}(W_{\mathrm{fuse},i,:}),
    \qquad
    f_t^{(i)}
    =
    \sum_{j=1}^{T}
    \alpha_j^{(i)}h_t^{(j)}.
    \label{eq:dflare-fusion}
\end{equation}
Because the fusion coefficients vary with draft depth, each layer receives a different view of the target hierarchy~\citep{zhang2026dflare}.
However, the scalar-weighted fusion is intentionally lightweight and does not retain the learned cross-layer transformation provided by DFlash's FC branch. Using it alone may therefore trade representational richness for layer specificity.

Motivated by their complementary roles, we formulate target conditioning as a shared-and-specialized decomposition.
The FC branch $c_t$ establishes a common semantic basis by modeling cross-layer feature interactions, while $f_t^{(i)}$ provides a depth-dependent refinement that preserves the target-layer preferences of draft layer $i$.
The resulting conditioning representation is
\begin{equation}
    g_t^{(i)}
    =
    \operatorname{RMSNorm}
    \left(
        c_t+f_t^{(i)}
    \right),
    \label{eq:dflarev2-hybrid-context}
\end{equation}
where the fusion is performed position-wise over the target context.
Draft layer $i$ uses the sequence of representations $g_t^{(i)}$ as its target context, retaining the globally transformed representation while receiving a specialized refinement aligned with its depth.
This decomposition avoids forcing global cross-layer interactions and layer-specific preferences into a single shared feature, thereby providing differentiated conditioning throughout the draft network.
The additional fusion branch introduces only $D\times T$ scalar weights; its softmax coefficients can be precomputed after training.
As shown later in Table~\ref{tab:drafter-ablation}, this hybrid backbone preserves DFlash's performance on math and chat while providing clear gains on code-generation benchmarks.

\begin{figure}[t]
    \centering
    \includegraphics[width=0.90\linewidth]{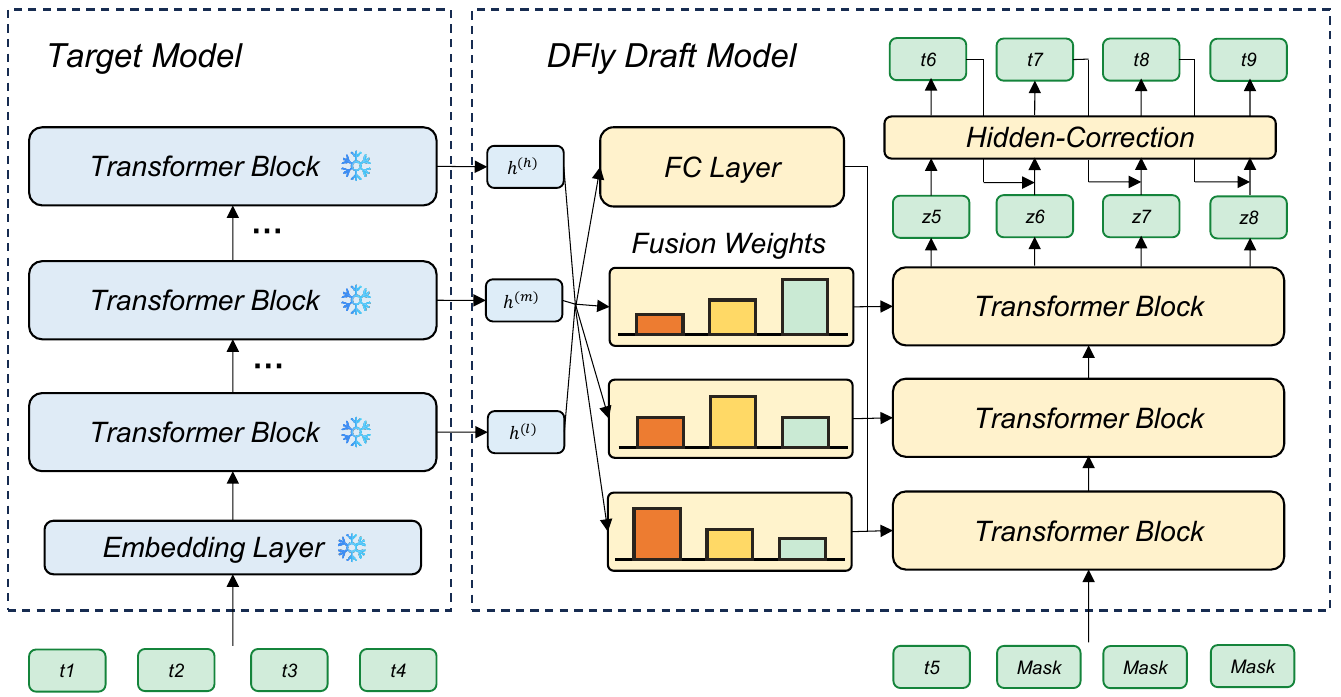}
    \caption{Overview of DFly. DFly uses a hybrid target-conditioning
backbone combines DFlash’s expressive shared projection with DFlare’s layer-specific target views. A lightweight autoregressive hidden-correction head introduces predecessor information after the parallel backbone.}
    \label{fig:placeholder}
\end{figure}

\subsection{Autoregressive Heads for Parallel Drafting}
Given an accepted context $x_{\leq t}$, the parallel backbone produces draft-model hidden states
$z^{\mathrm{D}}_{t+1},\ldots,z^{\mathrm{D}}_{t+B}$ and base logits
$U_{t+1},\ldots,U_{t+B}$ in one forward pass.
Here, the superscript $\mathrm{D}$ distinguishes states produced by the draft backbone from the target-model states $h_t^{(j)}$ used for conditioning above.
Although this makes drafting efficient, each base distribution approximates the marginal
$q(x_{t+i}\mid x_{\leq t})$ and cannot distinguish which continuation has actually been selected at earlier positions in the block.
We therefore place a lightweight sequential head after the parallel backbone.
It conditions each subsequent distribution on the realized draft prefix and induces the causal block distribution
\begin{equation}
    q(X_{t+1:t+B}\mid x_{\leq t})
    =
    \prod_{i=1}^{B}
    q_i(x_{t+i}\mid x_{\leq t},x_{t+1:t+i-1}).
    \label{eq:ar-factorization}
\end{equation}
The expensive backbone remains fully parallel; only this small head is evaluated from left to right.
On top of the hybrid backbone described above, we investigate two realizations of this autoregressive approximation.

\paragraph{Markov head.}
Following DSpark~\citep{dspark}, the first design adds a first-order transition bias to the parallel logits.
To avoid a full vocabulary-to-vocabulary transition matrix, it uses the low-rank factorization
$W_1W_2$, where $W_1\in\mathbb{R}^{V\times r}$ and $W_2\in\mathbb{R}^{r\times V}$:
\begin{equation}
    b_{t+i}=W_1[x_{t+i-1}]W_2,\qquad
    q_i=\operatorname{softmax}(U_{t+i}+b_{t+i}).
    \label{eq:dspark-markov}
\end{equation}
This head efficiently suppresses continuations that are incompatible with the sampled predecessor, but its transition is determined only by predecessor-token identity.

\paragraph{Hidden-correction head.}
Inspired by TreeFlash~\citep{rheinboldt2026treeflashparallelarapproximationfaster}, the second design introduces the predecessor token at the hidden-state level.
Let $\widetilde{z}^{\mathrm{D}}_{t+i}$ be the normalized draft-backbone state and $e_{t+i-1}$ the normalized embedding of the sampled predecessor.
We compute
\begin{equation}
    \widehat{z}^{\mathrm{D}}_{t+i}
    =
    z^{\mathrm{D}}_{t+i}
    +
    \operatorname{SwiGLU}
    \left([\widetilde{z}^{\mathrm{D}}_{t+i};e_{t+i-1}]\right),
    \qquad
    q_i=\operatorname{softmax}
    \left(\operatorname{LMHead}(\widehat{z}^{\mathrm{D}}_{t+i})\right).
    \label{eq:ar-approx}
\end{equation}
Unlike a fixed transition bias, this correction jointly uses the sampled predecessor and the position-specific representation produced by our hybrid backbone.
We adopt this hidden-correction head in \dflaretwo because it provides stronger context-dependent adaptation while adding only a small sequential module.

\subsection{Loss Design}
The acceptance-aligned objectives in \Cref{sec:mtp-loss} are used in two stages.
Although the end-to-end TV objective 
\cite{li2026breaking} in \Cref{eq:mtp-e2e} directly optimizes a surrogate of the expected accepted draft length, it is difficult to optimize from a poorly aligned initialization because the multiplicative acceptance terms can become very small at early training stages. We therefore first perform a cold-start stage with a D-PACE-weighted~\citep{wu2026d} hybrid LK loss \citep{lk-losses}, and then switch to the end-to-end TV objective for the final training stage.

During cold start, each per-depth hybrid LK loss from \Cref{eq:mtp-lk} is reweighted by a dynamic position-aware coefficient. 
For a training position $i$ and prediction depth $k$, let $s_{i,k}=q_{\psi}(x_{i+k+2})$ denote the draft confidence assigned to the supervised target token. 
In the implementation, this confidence is computed from the hard-label cross-entropy, i.e., $\operatorname{CE}_{i,k}=-\log q_{\psi}(x_{i+k+2})$ and $s_{i,k}=\exp(-\operatorname{CE}_{i,k})$.
We smooth the confidence as
\begin{equation}
    \tilde{s}_{i,k}
    =
    (1-\rho)s_{i,k}+\rho,
    \label{eq:mtp-dpace-smooth}
\end{equation}
where $\rho\in[0,1]$ is the D-PACE smoothing coefficient.
The D-PACE weight at depth $k$ is then computed as
\begin{equation}
    \bar{w}_{i,k}
    =
    \sum_{m=k}^{D-1}
    \prod_{r=0}^{m}
    \tilde{s}_{i,r}.
    \label{eq:mtp-dpace-weight}
\end{equation}
The cold-start objective is the weighted average
\begin{equation}
    \mathcal{L}_{\mathrm{cold}}
    =
    \frac{
    \sum_{k=0}^{D-1}
    \sum_{i\in\mathcal{I}_k}
    \operatorname{sg}[\bar{w}_{i,k}]
    \mathcal{L}_{\mathrm{LK}}^{\lambda}
    (p_{i,k},q_{i,k})
    }{
    \sum_{k=0}^{D-1}
    \sum_{i\in\mathcal{I}_k}
    \operatorname{sg}[\bar{w}_{i,k}]
    +
    \epsilon
    }.
    \label{eq:mtp-cold-dpace-lk}
\end{equation}
The D-PACE weights are detached from backpropagation and are used only to determine how much each depth contributes to the LK objective.

After the cold-start stage, we discard the D-PACE-weighted LK objective and continue training with the end-to-end TV objective in \Cref{eq:mtp-e2e}. This second stage directly optimizes the multi-depth acceptance surrogate rather than a per-position distributional objective.

In this way, the training schedule uses D-PACE-weighted LK loss to bring the drafter into a well-aligned regime, and then uses the end-to-end TV loss to optimize the final multi-token speculative acceptance behavior.










\subsection{Main Results}
\begin{table*}[t]
    \centering
    \small
    \setlength{\tabcolsep}{6pt}
    \renewcommand{\arraystretch}{1.2}
    \setlength{\aboverulesep}{0pt}
    \setlength{\belowrulesep}{0pt}
    \def\hlb{\cellcolor{HYLightBlue!55}}
    \def\hllb{\cellcolor{HYLightBlueLighter!55}}
    \caption{\textbf{Main draft-quality results on Qwen3-8B and Hy3-A21B at temperature 1, no thinking.}
    We report mean acceptance length for MTP, DFlash, and DSpark.
    Avg.\ is the arithmetic mean across the six reported benchmarks.
    \textbf{Bold} marks the best result in each column for each target model.}
    \label{tab:drafter-main-results}
    \resizebox{\textwidth}{!}{%
    \begin{tabular}{llccccccc}
        \toprule
        \multirow{2}{*}{\textbf{Target Model}}
          & \multirow{2}{*}{\textbf{Drafter}}
          & \multicolumn{2}{c}{\textbf{Math}}
          & \multicolumn{3}{c}{\textbf{Code}}
          & \multicolumn{1}{c}{\textbf{Chat}}
          & \multirow{2}{*}{\textbf{Avg.}} \\
        \cmidrule(lr){3-4}\cmidrule(lr){5-7}\cmidrule(lr){8-8}
          & & Math500 & GSM8K & HumanEval & MBPP & LiveCodeBench & MT-Bench & \\
        \midrule
        \multirow{4}{*}{Qwen3-8B}
 & MTP& 3.53& 3.56& 3.33& 3.22& 3.25& 2.57&3.24\\
          & DFlash & 4.97& 5.54& 4.77& 4.50& 4.46& 3.16& 4.57\\
          & DSpark & 5.87& 6.25& 5.56& 5.25& 5.20& \textbf{3.77}& 5.32\\
 & \hllb \dflaretwo & \hllb \textbf{6.06}& \hllb \textbf{6.42}& \hllb \textbf{5.60}& \hllb \textbf{5.34}& \hllb \textbf{5.36}& \hllb 3.67& \hllb\textbf{5.41}\\
        \midrule
        \multirow{3}{*}{\textbf{Hy3-A21B}}
          & MTP    & 3.30& 3.30& 3.13& 3.04& 2.84& 2.40& 3.00\\
          & DFlash & 4.01& 4.23& 4.36& 4.05& 3.10& 2.38& 3.69\\
          & \hlb \dflaretwo & \hlb \textbf{5.23}& \hlb \textbf{5.53}& \hlb \textbf{5.52}& \hlb \textbf{5.41}& \hlb \textbf{4.07}& \hlb \textbf{2.96}& \hlb \textbf{4.79}\\
        \bottomrule
    \end{tabular}%
    }
\end{table*}

\paragraph{Setup.}

We evaluate \dflaretwo on two target models, Qwen3-8B~\citep{qwen3technicalreport} and Hy3-A21B, and compare it with three representative drafters: MTP~\citep{mtp}, a multi-token prediction baseline; DFlash~\citep{dflash}, a fully parallel block drafter; and DSpark~\citep{dspark}, a semi-autoregressive parallel drafter.
We align MTP's TTT horizon and the MTP prediction horizon with the eight-token block used by DFlash, DSpark, and \dflaretwo.
MTP uses one draft Transformer layer, whereas DFlash, DSpark, and \dflaretwo use five draft layers.
For Qwen3-8B, we take prompts from Open-PerfectBlend~\citep{xu2024perfect}, regenerate responses with that target in non-thinking mode, and train every drafter for ten epochs with the same learning-rate schedule for fair comparison. For Hy3-A21B, we take the best MTP model described in Section~\ref{sec:mtp} and compared it with our DFlash and \dflaretwo models strengthened with code- and mathematics-focused samples described in Section~\ref{sec:dfly_ablation}.   We evaluate tasks in three categories: Math: GSM8K~\citep{cobbe2021gsm8k}, MATH500~\citep{hendrycks2021MATH}; Code: HumanEval~\citep{chen2021humaneval}, MBPP~\citep{austin2021mbpp}, LiveCodeBench~\citep{jain2024livecodebench}; Chat: MTBench~\citep{zheng2023mtbench}. We only report  mean acceptance length to isolate draft quality; Section~\ref{sec:dcut} evaluates end-to-end throughput for these methods under different serving concurrencies.

\paragraph{Main Results.} 
\label{sec:drafter-ablation}

Table~\ref{tab:drafter-main-results} compares the mean acceptance length of representative drafters on Qwen3-8B and Hy3-A21B.
On Qwen3-8B, \dflaretwo achieves the best result on all five math and code benchmarks and raises the overall average to $5.41$, compared with $5.32$ for DSpark, $4.57$ for DFlash, and $3.24$ for MTP.
Its advantage is most visible on structured generation: it reaches $6.06$ on Math500, $6.42$ on GSM8K, and improves all three code benchmarks over DSpark.
DSpark remains slightly better on MT-Bench, consistent with our use of \dflaretwo primarily for code- and math-oriented workloads. The improvement is larger on Hy3-A21B.
\dflaretwo outperforms both MTP and DFlash on every reported benchmark, increasing the overall average acceptance length to $4.79$, compared with $3.00$ for MTP and $3.69$ for DFlash.
This corresponds to relative gains of $59.7\%$ over MTP and $29.8\%$ over DFlash.
The consistent improvement across all six benchmarks shows that the hybrid target-conditioning backbone and autoregressive correction transfer effectively to Hy3-A21B, with particularly strong benefits on structured code and mathematical reasoning.

\begin{table*}[t]
    \centering
    \small
    \setlength{\tabcolsep}{6pt}
    \renewcommand{\arraystretch}{1.2}
    \setlength{\aboverulesep}{0pt}
    \setlength{\belowrulesep}{0pt}
    \caption{\textbf{Ablation of the parallel-drafting architecture on Hy3 under greedy decoding (\(T=0\), no thinking).}
    We report mean acceptance length; all variants use D-PACE.
    Avg.\ is the arithmetic mean across the six reported benchmarks.
    The comparison is cumulative: both AR heads are added to the selected \dflaretwo backbone, and data expansion is applied to the hidden-correction variant.
    \textbf{Bold} marks the best result in each column.}
    \label{tab:drafter-ablation}
    \resizebox{\textwidth}{!}{%
    \begin{tabular}{llccccccc}
        \toprule
        \multirow{2}{*}{\textbf{Component}}
          & \multirow{2}{*}{\textbf{Config}}
          & \multicolumn{2}{c}{\textbf{Math}}
          & \multicolumn{3}{c}{\textbf{Code}}
          & \multicolumn{1}{c}{\textbf{Chat}}
          & \multirow{2}{*}{\textbf{Avg.}} \\
        \cmidrule(lr){3-4}\cmidrule(lr){5-7}\cmidrule(lr){8-8}
          & & Math500 & GSM8K & HumanEval & MBPP & LiveCodeBench & MT-Bench & \\
        \midrule
        \multirow{2}{*}{Backbone}
          & DFlash                         & 4.27& 3.86& 4.27& 4.18& 3.46& 2.60& 3.77\\
          & \dflaretwo                  & 4.92& 4.88& 5.00& 5.03& 3.71& 2.88& 4.40\\
        \midrule
        \multirow{2}{*}{\(+\) AR head}
          & Markov                         & 5.12& 5.13& 5.24& 4.98& 3.85& 3.03& 4.56\\
          & Hidden             & 5.13& 5.17& 5.26& 5.00& 4.00& \textbf{3.06}& 4.60\\
        \midrule
        \(+\) DATA
          & Code Math& \textbf{5.33}& \textbf{5.18}& \textbf{5.53}&  \textbf{5.45}& \textbf{4.02}& 3.02& \textbf{4.75}\\
        \bottomrule
    \end{tabular}%
    }
\end{table*}

\subsection{Ablation Study}
\label{sec:dfly_ablation}
To disentangle the contribution of each component, we conduct detailed ablation studies on Hy3-A21B.
Unless otherwise specified, all variants are trained for six epochs on Open-PerfectBlend responses regenerated by the target model, using the D-PACE-weighted LK objective.
Our analysis follows the construction of the final model: we first compare alternative target-conditioning backbones, then evaluate different autoregressive heads on the selected \dflaretwo backbone, and finally measure the effect of domain-specific data expansion on the complete \dflaretwo architecture.
We additionally ablate the  draft-model depth, and inference block size and thinking mode to determine the loss and model configuration used in our final system.

\paragraph{Backbone.}
Table~\ref{tab:drafter-ablation} first isolates the target-conditioning backbone.
Compared with DFlash, \dflaretwo remains comparable on math and chat.
Its advantage is more pronounced on code, for example, improving HumanEval from $4.27$ to $5.00$.
This domain pattern supports our design goal: the shared nonlinear context preserves the general capability of DFlash, while layer-specific target views provide additional capacity for structured code continuations.
We therefore select \dflaretwo as the default backbone for code- and math-oriented drafting.

\paragraph{Autoregressive head.}
The parallel backbone predicts each block position from the accepted context but does not observe the continuation selected at earlier draft positions.
The sequential head addresses this mismatch by converting marginal parallel predictions into distributions conditioned on the realized predecessor.
On the same \dflaretwo backbone, the Markov head raises the overall average acceptance length from $4.40$ to $4.56$, while hidden correction further improves it to $4.60$.
Hidden correction also outperforms the Markov head on every reported benchmark.
Conditioning the correction jointly on the predecessor embedding and position-specific hidden state is therefore more effective than applying a fixed token-transition bias, motivating our choice of the hidden-correction head.
Section~\ref{sec:modular_analysis} further analyzes its runtime overhead and shows that the additional cost is covered by the increase in accepted length.

\paragraph{Domain-specific data expansion.}
To further strengthen code generation and mathematical reasoning, we augment Open-PerfectBlend with $700$K domain-specific prompts.
For code, we select $500$K prompts in total from OpenCodeInstruct~\citep{ahmad2025opencodeinstruct} and OpenCodeReasoning~\citep{ahmad2025opencodereasoning}.
The former provides broad code instruction coverage, while the latter emphasizes distilled reasoning traces for competitive programming; together, they improve both general code completion and multi-step program synthesis.
For mathematics, we sample another $200$K prompts from Big-Math~\citep{albalak2025bigmathlargescalehighqualitymath}, providing additional coverage of diverse and challenging reasoning problems.
Before response generation, we remove any prompt that shares a contiguous sequence of at least $16$ tokens with an evaluation example.
We then regenerate responses for the filtered prompts with Hy3-A21B, mix the resulting code and math data with Open-PerfectBlend, and train the combined corpus for six epochs using the same objective.
As shown in Table~\ref{tab:drafter-ablation}, this expansion increases the math/code average acceptance length largely.
The gains are consistent across all structured benchmarks, except MT-Bench decreases slightly from $3.06$ to $3.02$, reflecting the intended specialization toward code and math rather than a uniform shift across domains.
The resulting checkpoint is used as our final code/math drafter.

\paragraph{Draft depth and inference block size.}
We further ablate the draft depth and the inference block size while keeping the target-side configuration fixed.
This ablation is motivated by the latency behavior in \Cref{fig:hy_component_latency}c.
At comparable nominal effective concurrency, the average decode-step latency proxy from draft generation to target verification is similar for the 5-layer and 7-layer drafters.
Therefore, in this regime, using a deeper drafter does not substantially change the per-iteration execution cost; the final efficiency is instead mainly determined by how many tokens are accepted in each speculative iteration.
As shown in \Cref{tab:draft-depth-block-ablation}, the 3-layer drafter with block size 5 achieves a lower mean accepted length than the 5-layer drafter, indicating that an overly shallow drafter weakens the quality of speculative proposals.
However, once the draft depth is increased to 5 layers, the inference block size becomes the more influential factor.
Increasing the block size from 5 to 8 improves the average accepted length from 3.39 to 4.25.
This indicates that exposing more draft positions to the target model is particularly beneficial under the $T=1$ no-thinking setting, especially on math and code benchmarks.
Although using a smaller block size reduces the nominal effective concurrency and hence lowers the overall per-iteration latency, it also shortens the accepted length, reducing the amount of useful progress made in each speculative step.
When the block size is fixed to 8, further increasing the draft depth from 5 to 7 layers provides almost no additional average improvement, with both configurations achieving an average accepted length of 4.25.
Together with the latency trend in \Cref{fig:hy_component_latency}c, this suggests that the 5-layer drafter with block size 8 offers a better efficiency--performance trade-off: it matches the best average accepted length while avoiding the additional depth of the 7-layer drafter.
Based on this observation, we use the 5-layer drafter with block size 8 as the default configuration unless otherwise specified.

\begin{table*}[t]
    \centering
    \small
    \setlength{\tabcolsep}{6pt}
    \renewcommand{\arraystretch}{1.2}
    \setlength{\aboverulesep}{0pt}
    \setlength{\belowrulesep}{0pt}
    \caption{\textbf{Draft depth and inference block-size comparison under the $T=1$ no-thinking setting.}
    Each entry is the mean accepted length under the fixed target-side configuration.
    Avg.\ is computed over the six displayed benchmarks.
    \textbf{Bold} marks the best result in each column.}
    \label{tab:draft-depth-block-ablation}
    \resizebox{\textwidth}{!}{%
    \begin{tabular}{ccccccccc}
        \toprule
        \multirow{2}{*}{\textbf{Draft layers}}
          & \multirow{2}{*}{\textbf{Block size}}
          & \multicolumn{2}{c}{\textbf{Math}}
          & \multicolumn{3}{c}{\textbf{Code}}
          & \multicolumn{1}{c}{\textbf{Chat}}
          & \multirow{2}{*}{\textbf{Avg.}} \\
        \cmidrule(lr){3-4}\cmidrule(lr){5-7}\cmidrule(lr){8-8}
          & & Math500 & GSM8K & HumanEval & MBPP & LiveCodeBench & MT-Bench & \\
        \midrule
        3 & 5
          & 3.53& 3.72& 3.76& 3.52& 2.86& 2.32& 3.29\\
        5 & 5
          & 3.64& 3.84& 3.79& 3.64& 3.03& 2.42& 3.39\\
        5 & 8
          & \textbf{4.66}& \textbf{5.10}& \textbf{5.02}& 4.58& 3.49& 2.66& 4.25\\
        7 & 8
          & 4.64& 5.05& 4.90& \textbf{4.59}& \textbf{3.63}& \textbf{2.67}& \textbf{4.25}\\
        \bottomrule
    \end{tabular}%
    }
\end{table*}

\begin{table*}[t]
    \centering
    \small
    \def\hlb{\cellcolor{HYLightBlue!55}}
    \setlength{\tabcolsep}{6pt}
    \renewcommand{\arraystretch}{1.2}
    \setlength{\aboverulesep}{0pt}
    \setlength{\belowrulesep}{0pt}
    \caption{\textbf{Thinking-mode ablation of HY3 DeepSpec under the $T=1$ setting.}
    Each entry is the mean accepted length under the corresponding training and evaluation thinking-mode configuration.
    Avg.\ is computed over the six displayed benchmarks when all entries are available.
    \textbf{Bold} marks the best result in each column among complete configurations.}
    \label{tab:hy3-deepspec-thinking-mode}
    \resizebox{\textwidth}{!}{%
    \begin{tabular}{llccccccc}
        \toprule
        \multirow{2}{*}{\textbf{Train Data}}
          & \multirow{2}{*}{\textbf{Test Data}}
          & \multicolumn{2}{c}{\textbf{Math}}
          & \multicolumn{3}{c}{\textbf{Code}}
          & \multicolumn{1}{c}{\textbf{Chat}}
          & \multirow{2}{*}{\textbf{Avg.}} \\
        \cmidrule(lr){3-4}
        \cmidrule(lr){5-7}
        \cmidrule(lr){8-8}
          & & Math500 & GSM8K & HumanEval & MBPP
          & LiveCodeBench & MT-Bench & \\
        \midrule
        No-think& No-think& \hlb \textbf{5.23}& \hlb \textbf{5.53}& \hlb \textbf{5.52}& \hlb \textbf{5.41}& \hlb \textbf{4.07}& \hlb 2.96& \hlb \textbf{4.79}\\
        No-think
          & High-think
          & 3.30
          & 3.27
          & 3.60
          & 3.55
          & 3.18
          & 2.82
          & 3.29 \\
           \midrule
        High-think
          & High-think
          & \hlb 4.29
          & \hlb 4.47
          & \hlb 4.76
          & \hlb 4.74
          & \hlb 3.92
          & \hlb \textbf{3.42}
          & \hlb 4.27 \\
        High-think
          & No-think
          &  4.93
          &  5.30
          &  5.30
          & 4.87
          & 4.21& 3.04
          & 4.61\\
        \bottomrule
    \end{tabular}%
    }
\end{table*}

\paragraph{Thinking-Mode Specialization.}
Table~\ref{tab:hy3-deepspec-thinking-mode} compares different Hy3 training and inference thinking-mode configurations under the $T=1$ setting.
The results show that the best accepted length is achieved only when the draft model is trained and evaluated under the matched thinking mode.
Specifically, the no-thinking draft model obtains the highest overall average when evaluated with no-thinking prompts, reaching 4.79 across the six benchmarks.
However, when the same no-thinking draft model is directly evaluated under the high-thinking setting, the average accepted length drops significantly to 3.29.
Conversely, the high-thinking draft model performs best when evaluated under the high-thinking setting, achieving stronger results on reasoning- and chat-oriented workloads, especially on MT-Bench where it reaches 3.42.
These results indicate that high-think and no-think decoding induce different hidden-state and acceptance distributions.
Therefore, high-think and no-think draft models should be trained separately and deployed with their corresponding target-side thinking mode, rather than sharing a single draft model across both inference modes.
\section{Real-World Deployment: D-cut}

\label{sec:dcut}

The drafters presented in Sections~\ref{sec:mtp}--\ref{sec:dflare} improve proposal quality and extend the useful draft horizon from the training and architecture perspectives.
In deployment, however, throughput depends on both the number of tokens committed per speculative step and the wall-clock time of that step.
A longer draft can improve the former while degrading the latter, because every retained position must still be evaluated by the target model.
\dcut~\citep{dcut} addresses this serving-side trade-off for long-block drafting.
At every decoding step, it answers two coupled questions: how should a verification budget be distributed across requests, and how large should that budget be under the current runtime condition?

\subsection{The Deployment Challenge}

\label{sec:dcut-challenge}

Consider a batch of \(B\) active requests.
For request \(i\), the drafter proposes \(D\) tokens \(z_{i,1:D}\); together with the certain bonus token \(z_{i,0}\), full-depth verification processes \(B(D{+}1)\) positions.
If only the first \(n_i\in\{0,\ldots,D\}\) drafts are retained, request \(i\) contributes \(n_i{+}1\) verification positions, and the total batch budget is \(K=\sum_{i=1}^{B}(n_i{+}1)\).
Choosing the keep depths \(n_{1:B}\) is difficult for two reasons.

\paragraph{Useful depth varies across requests.}
The acceptance length changes across requests, domains, and decoding steps.
A uniform keep depth therefore spends the same target-model compute on requests with long, high-confidence continuations and requests whose drafts are likely to fail after only a few positions.
The latter contribute little progress but can dominate the verification workload.

\paragraph{The cost of depth varies across deployments.}
The latency added by another verification position depends on the active batch size, GPU, model architecture, parallelism configuration, and CUDA-graph execution path.
When verification is underutilized, retaining additional positions is nearly free; once it becomes compute-bound, the same positions can substantially increase step latency.
Consequently, neither one uniform per-request depth nor one fixed global budget is suitable for all serving conditions.

\subsection{D-cut: Adaptive Verification Budgeting}
\label{sec:dcut-method}

\begin{figure}[t]
    \centering
    \includegraphics[width=\textwidth]{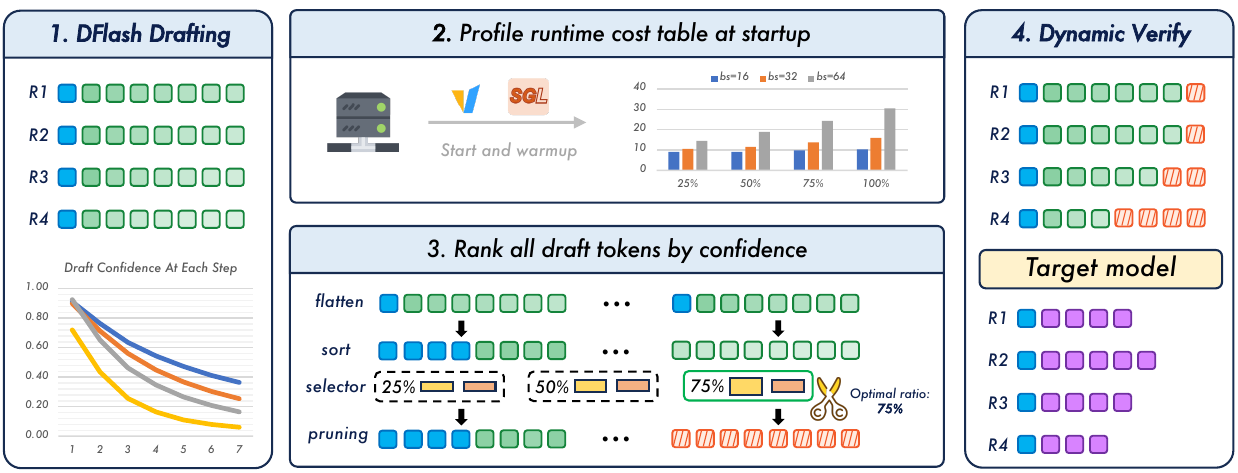}
    \caption{Overview of \dcut. The drafter produces batched draft blocks with token-level confidence scores.
\dcut evaluates the expected utility of each position, dynamically selects a runtime-aware global budget, and
packs only the top candidates into a dense verification batch for the target model.}
\label{fig:dcut-method}
\end{figure}

\dcut treats target-model verification as a shared batch-level resource.
It first estimates the benefit of each possible keep depth, then combines the resulting batch benefit with a profiled runtime cost to select the operating point.

\paragraph{Estimating per-request progress.}
Let \(L_i\in\{0,\ldots,D\}\) denote the acceptance length that would be observed if request \(i\)'s full block were verified.
Keeping \(n_i\) drafts advances \(A_i(n_i)=1+\min(L_i,n_i)\) tokens, including the certain bonus token.
Its expectation can be written in terms of prefix-survival probabilities:
\begin{equation}
    \mathbb{E}[A_i(n_i)]
    = 1 + \sum_{k=1}^{n_i}\Pr(L_i\ge k).
    \label{eq:dcut-expected-progress}
\end{equation}
The true survival probabilities are unknown before verification, so \dcut estimates them from the drafter's token-level confidence \(c_{i,t}\):
\begin{equation}
    s_{i,k} = \prod_{t=1}^{k} c_{i,t}, \quad s_{i,0}=1,
    \label{eq:dcut-score}
\end{equation}
where the prefix product reflects that position \(k\) is useful only if all preceding positions survive.
Substituting \(s_{i,k}\) for \(\Pr(L_i\ge k)\) in \Cref{eq:dcut-expected-progress} gives the estimated progress
\begin{equation}
    \hat{A}_i(n_i) = \sum_{k=0}^{n_i} s_{i,k}.
    \label{eq:dcut-expected-advance}
\end{equation}
Because \(s_{i,k}\) is non-increasing with \(k\), selecting the top-\(K\) scores across all requests, with ties broken toward shallower positions, retains a contiguous prefix for each request.
This converts a global budget into request-specific keep depths without producing invalid holes inside a draft block.

\paragraph{Selecting a runtime-aware global budget.}
\dcut restricts the budget to a small set of discrete ratios \(\mathcal{R}{=}\{0.25, 0.50, 0.75, 1.00\}\).
For ratio \(\rho\) and batch size \(B\), the budget is:
\begin{equation}
    K_\rho(B) = \max\!\left(B,\; \left\lceil \rho \cdot B(D{+}1) \right\rceil\right),
    \label{eq:dcut-budget}
\end{equation}
where the lower bound \(B\) ensures every request retains at least its bonus token.
For each candidate ratio, top-\(K_\rho(B)\) selection induces keep depths \(n_i^{(\rho)}\).
We define the corresponding estimated batch progress as
\begin{equation}
    U(B,\rho)
    = \sum_{i=1}^{B}\hat{A}_i\!\left(n_i^{(\rho)}\right).
    \label{eq:dcut-batch-utility}
\end{equation}
At server startup, \dcut profiles the end-to-end speculative-step latency \(C(B,\rho)\) for each supported \((B,\rho)\) pair, including drafting, selection, packing, and target verification.
At runtime, it chooses
\begin{equation}
    \rho^\star
    = \arg\max_{\rho\in\mathcal{R}}
      \frac{U(B,\rho)}{C(B,\rho)}.
    \label{eq:dcut-select}
\end{equation}
For a fixed \(B\), this objective maximizes projected output-token throughput.
The AR step time and the normalization by \(B\) are constant across candidate ratios, so they do not affect the ratio selected when the objective is expressed as projected speedup over AR.

\paragraph{Packing and correctness.}
After selecting \(\rho^\star\), \dcut packs the unequal request prefixes into one dense verification batch of \(K_{\rho^\star}(B)\) positions and submits it to the target model.
It changes only how far each draft is evaluated; retained tokens still pass through the original target-model verification.
Under the deterministic-draft and target-only verification protocol used in our deployment experiments, this preserves the target-model output distribution exactly.
\dcut therefore trades some expected progress per step for a shorter step time; it does not guarantee a throughput gain for every batch, which is why both benefit and cost appear in \Cref{eq:dcut-select}.
Further algorithmic and correctness details are provided in the companion paper~\citep{dcut}.

\subsection{Role within AngelSpec}
\label{sec:dcut-interaction}

\dcut is orthogonal to drafter training and model architecture: it consumes proposed blocks and confidence scores but does not alter either model's weights.
Within \method, it is primarily intended for the block-parallel drafting path.
The default \dflaretwo configuration of \Cref{sec:dflare} proposes \(D{=}8\) draft tokens, so each request contributes eight block positions per step and the block suffix carries most of the low-utility mass.
The short-horizon MTP path uses \(D{=}3\), leaving a smaller verification surface and correspondingly less room to prune.
The mechanism applies to any batched drafter that exposes per-position confidence, but the results below evaluate only the \dflaretwo path.

Selecting \(\rho{=}1\) lets \dcut retain the full block when pruning is not worthwhile, but the selector and packing logic still introduce overhead.
Moreover, the current implementation supports only piecewise CUDA graphs.
Thus, \dcut is not assumed to be free: the profiled end-to-end cost in \Cref{eq:dcut-select} is essential to deciding whether a smaller verification budget is beneficial.

Both constraints are implementation limits rather than properties of the method, and we are actively working to remove them.
Because \(\mathcal{R}\) contains only a few ratios, the reachable verification shapes form a small finite set, so full CUDA graph capture is feasible once a graph is captured per \((B,\rho)\) pair.
We are likewise moving budget selection off the critical path so that it overlaps target-model execution under asynchronous scheduling, instead of forming a synchronization point between drafting and verification.
Both changes lower the cost term \(C(B,\rho)\) without altering the selection rule, so they should widen the regime in which retaining a longer draft remains profitable.

\subsection{Deployment Results}
\label{sec:dcut-results}

\begin{table*}[t]
    \centering
    \small
    \setlength{\tabcolsep}{4.5pt}
    \renewcommand{\arraystretch}{1.2}
    \setlength{\aboverulesep}{0pt}
    \setlength{\belowrulesep}{0pt}
    \def\hlb{\cellcolor{HYLightBlue!55}}
    \caption{\textbf{HY3 (295B-A21B, TP=8) throughput at temperature 1 across concurrency levels.} Tok/s: output-token throughput. Spd.: throughput relative to AR. \textbf{bold} marks the best speedup for each dataset at each concurrency. Every cell uses \(3\times120\)\,s windows; Avg.\ is the arithmetic mean across six datasets.}
    \label{tab:dfly-hy3}
    \resizebox{\textwidth}{!}{%
    \begin{tabular}{cl*{7}{rc}}
        \toprule
        & & \multicolumn{2}{c}{\textbf{GSM8K}}
          & \multicolumn{2}{c}{\textbf{Math500}}
          & \multicolumn{2}{c}{\textbf{HumanEval}}
          & \multicolumn{2}{c}{\textbf{MBPP}}
          & \multicolumn{2}{c}{\textbf{LiveCodeBench}}
          & \multicolumn{2}{c}{\textbf{MT-Bench}}
          & \multicolumn{2}{c}{\textbf{Avg.}} \\
        \cmidrule(lr){3-4}\cmidrule(lr){5-6}\cmidrule(lr){7-8}\cmidrule(lr){9-10}\cmidrule(lr){11-12}\cmidrule(lr){13-14}\cmidrule(lr){15-16}
        \textbf{Conc.} & \textbf{Method}
          & Tok/s & Spd.
          & Tok/s & Spd.
          & Tok/s & Spd.
          & Tok/s & Spd.
          & Tok/s & Spd.
          & Tok/s & Spd.
          & Tok/s & Spd. \\
        \midrule
        \multirow{4}{*}{c4}
          & AR                    & 287.9 & 1.00$\times$ & 293.9 & 1.00$\times$ & 279.7 & 1.00$\times$ & 294.4 & 1.00$\times$ & 284.5 & 1.00$\times$ & 290.0 & 1.00$\times$ & 288.4 & 1.00$\times$ \\
          & MTP-3                 & 495.5 & 1.72$\times$ & 517.0 & 1.76$\times$ & 473.6 & 1.69$\times$ & 476.7 & 1.62$\times$ & 414.0 & 1.46$\times$ & 384.5 & \textbf{1.33$\times$} & 460.2 & 1.60$\times$ \\
          & DFlash-8              & 569.5 & 1.98$\times$ & 587.9 & 2.00$\times$ & 588.1 & 2.10$\times$ & 575.7 & 1.96$\times$ & 408.4 & 1.44$\times$ & 371.6 & 1.28$\times$ & 516.9 & 1.79$\times$ \\
          & \hlb DFly-8           & \hlb 635.4 & \hlb \textbf{2.21$\times$} & \hlb 643.9 & \hlb \textbf{2.19$\times$} & \hlb 647.2 & \hlb \textbf{2.31$\times$} & \hlb 661.5 & \hlb \textbf{2.25$\times$} & \hlb 455.2 & \hlb \textbf{1.60$\times$} & \hlb 384.0 & \hlb 1.32$\times$ & \hlb 571.2 & \hlb \textbf{1.98$\times$} \\
        \midrule
        \multirow{4}{*}{c8}
          & AR                    & 426.3 & 1.00$\times$ & 435.7 & 1.00$\times$ & 400.3 & 1.00$\times$ & 433.7 & 1.00$\times$ & 413.4 & 1.00$\times$ & 421.8 & 1.00$\times$ & 421.8 & 1.00$\times$ \\
          & MTP-3                 & 756.9 & 1.78$\times$ & 791.5 & 1.82$\times$ & 717.4 & 1.79$\times$ & 729.1 & 1.68$\times$ & 620.8 & 1.50$\times$ & 590.1 & \textbf{1.40$\times$} & 701.0 & 1.66$\times$ \\
          & DFlash-8              & 857.0 & 2.01$\times$ & 860.1 & 1.97$\times$ & 866.2 & 2.16$\times$ & 866.2 & 2.00$\times$ & 609.1 & 1.47$\times$ & 545.5 & 1.29$\times$ & 767.4 & 1.82$\times$ \\
          & \hlb DFly-8           & \hlb 964.8 & \hlb \textbf{2.26$\times$} & \hlb 959.5 & \hlb \textbf{2.20$\times$} & \hlb 974.1 & \hlb \textbf{2.43$\times$} & \hlb 1001.5 & \hlb \textbf{2.31$\times$} & \hlb 641.3 & \hlb \textbf{1.55$\times$} & \hlb 565.8 & \hlb 1.34$\times$ & \hlb 851.2 & \hlb \textbf{2.02$\times$} \\
        \midrule
        \multirow{4}{*}{c16}
          & AR                    & 650.7 & 1.00$\times$ & 670.6 & 1.00$\times$ & 594.8 & 1.00$\times$ & 667.9 & 1.00$\times$ & 617.2 & 1.00$\times$ & 628.1 & 1.00$\times$ & 638.2 & 1.00$\times$ \\
          & MTP-3                 & 1146.6 & 1.76$\times$ & 1174.0 & 1.75$\times$ & 1076.1 & 1.81$\times$ & 1103.3 & 1.65$\times$ & 893.5 & 1.45$\times$ & 876.6 & 1.40$\times$ & 1045.0 & 1.64$\times$ \\
          & DFlash-8              & 1446.1 & 2.22$\times$ & 1430.4 & 2.13$\times$ & 1433.3 & 2.41$\times$ & 1440.4 & 2.16$\times$ & 970.5 & 1.57$\times$ & 901.0 & 1.43$\times$ & 1270.3 & 1.99$\times$ \\
          & \hlb DFly-8           & \hlb 1623.8 & \hlb \textbf{2.50$\times$} & \hlb 1607.5 & \hlb \textbf{2.40$\times$} & \hlb 1595.4 & \hlb \textbf{2.68$\times$} & \hlb 1677.8 & \hlb \textbf{2.51$\times$} & \hlb 1046.7 & \hlb \textbf{1.70$\times$} & \hlb 961.5 & \hlb \textbf{1.53$\times$} & \hlb 1418.8 & \hlb \textbf{2.22$\times$} \\
        \midrule
        \multirow{4}{*}{c32}
          & AR                    & 918.6 & 1.00$\times$ & 1000.8 & 1.00$\times$ & 850.8 & 1.00$\times$ & 1018.7 & 1.00$\times$ & 893.2 & 1.00$\times$ & 921.1 & 1.00$\times$ & 933.9 & 1.00$\times$ \\
          & MTP-3                 & 1923.8 & 2.09$\times$ & 1989.8 & 1.99$\times$ & 1780.7 & 2.09$\times$ & 1863.6 & 1.83$\times$ & 1394.2 & 1.56$\times$ & 1469.4 & 1.60$\times$ & 1736.9 & 1.86$\times$ \\
          & DFlash-8              & 2261.8 & 2.46$\times$ & 2334.3 & 2.33$\times$ & 2170.6 & 2.55$\times$ & 2339.3 & 2.30$\times$ & 1489.9 & 1.67$\times$ & 1453.4 & 1.58$\times$ & 2008.2 & 2.15$\times$ \\
          & \hlb DFly-8           & \hlb 2527.4 & \hlb \textbf{2.75$\times$} & \hlb 2608.9 & \hlb \textbf{2.61$\times$} & \hlb 2429.1 & \hlb \textbf{2.86$\times$} & \hlb 2741.3 & \hlb \textbf{2.69$\times$} & \hlb 1602.6 & \hlb \textbf{1.79$\times$} & \hlb 1549.4 & \hlb \textbf{1.68$\times$} & \hlb 2243.1 & \hlb \textbf{2.40$\times$} \\
        \midrule
        \multirow{4}{*}{c64}
          & AR                    & 1156.9 & 1.00$\times$ & 1381.3 & 1.00$\times$ & 1197.6 & 1.00$\times$ & 1513.9 & 1.00$\times$ & 1229.6 & 1.00$\times$ & 1306.9 & 1.00$\times$ & 1297.7 & 1.00$\times$ \\
          & MTP-3                 & 2932.0 & \textbf{2.53$\times$} & 3170.0 & 2.29$\times$ & 2639.6 & 2.20$\times$ & 3011.3 & 1.99$\times$ & 2050.4 & 1.67$\times$ & 2350.0 & \textbf{1.80$\times$} & 2692.2 & 2.08$\times$ \\
          & DFlash-8              & 2523.4 & 2.18$\times$ & 2947.1 & 2.13$\times$ & 2655.2 & 2.22$\times$ & 2671.8 & 1.76$\times$ & 2015.8 & 1.64$\times$ & 1815.8 & 1.39$\times$ & 2438.2 & 1.89$\times$ \\
          & \hlb DFly-8           & \hlb 2827.8 & \hlb 2.44$\times$ & \hlb 3301.8 & \hlb \textbf{2.39$\times$} & \hlb 2965.0 & \hlb \textbf{2.48$\times$} & \hlb 3130.9 & \hlb \textbf{2.07$\times$} & \hlb 2195.0 & \hlb \textbf{1.79$\times$} & \hlb 1936.6 & \hlb 1.48$\times$ & \hlb 2726.2 & \hlb \textbf{2.11$\times$} \\
        \bottomrule
    \end{tabular}%
    }
\end{table*}

\Cref{tab:dfly-hy3} reports the drafter-level comparison on the benchmark sweep.
\dflaretwo attains the highest average speedup at every concurrency level and leads on the code- and mathematics-oriented benchmarks, while MTP-3 remains preferable on MT-Bench, consistent with the workload split motivating \method.
We therefore take \dflaretwo as the baseline that \dcut must improve upon.

\begin{figure}[t]
    \centering
    \includegraphics[width=\textwidth]{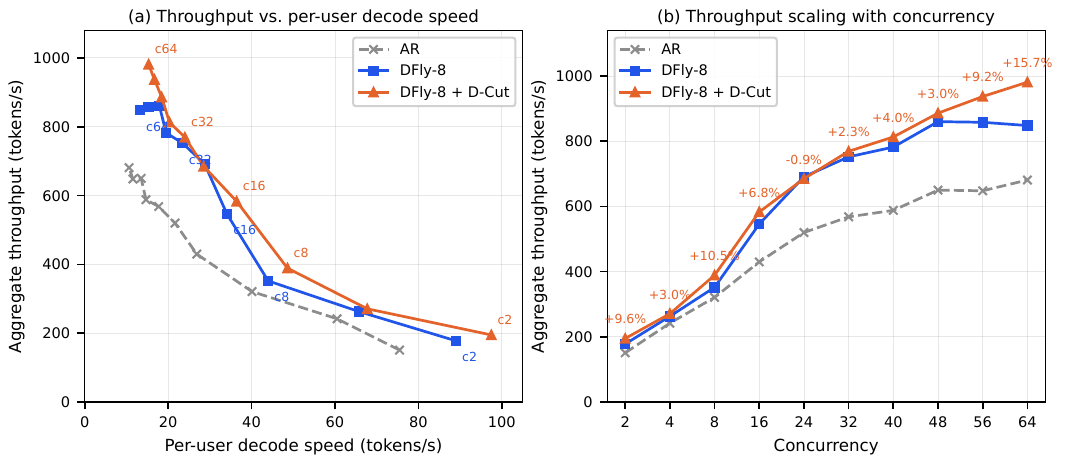}
    \caption{\textbf{\dcut on Hunyuan live traffic} (Hy3-295B-A21B, TP=8, \(8\times\) H20; concurrency 2--64).
    (a) Aggregate throughput against per-user decode speed, the frontier a deployment actually operates on; points up and to the right are better, and selected concurrency levels are annotated.
    (b) Aggregate throughput against concurrency, with labels giving \dcut's throughput relative to \dflaretwo at each point.
    \dflaretwo saturates beyond concurrency 48, whereas \dcut continues to convert additional load into throughput.
    \dflaretwo uses full-and-piecewise CUDA graph capture and \dcut piecewise capture only, so the comparison understates \dcut.}
    \label{fig:dcut-live}
\end{figure}

To measure \dcut under realistic serving conditions rather than a fixed benchmark mixture, we replay Hunyuan production traffic against Hy3-295B-A21B with TP=8 on \(8\times\) H20 GPUs, following the same measurement protocol as \Cref{tab:dfly-hy3} and sweeping concurrency from 2 to 64.
Live traffic is the relevant test for \dcut because its budget decision reacts to the request mix and load actually present at each step, which a per-dataset benchmark holds artificially constant.
\Cref{fig:dcut-live} reports aggregate throughput, per-user decode speed, and acceptance length.

\paragraph{\dcut extends the useful concurrency range.}
\dflaretwo saturates past concurrency 48, reaching 860, 858, and 848 tok/s at c48, c56, and c64.
Beyond that point additional concurrency buys no aggregate throughput while per-user decode speed keeps degrading, which is the signature of verification-bound serving.
\dcut converts the same load into throughput over the same range, reaching 886, 937, and 981 tok/s, or \(+3.0\%\), \(+9.2\%\), and \(+15.7\%\) relative to \dflaretwo.

\paragraph{A better latency--throughput frontier.}
Panel~(a) shows the consequence for deployment: the two curves separate exactly where operators care, in the high-load region.
At a matched per-user decode speed of roughly \(15.3\) tok/s, \dflaretwo sustains 858 tok/s at c56 while \dcut sustains 981 tok/s at c64, \(14\%\) higher aggregate throughput at the same per-user latency.
Measured against AR, \dflaretwo peaks near \(1.33\times\) between c24 and c40 and then falls back to \(1.25\times\) at c64, whereas \dcut keeps improving to \(1.45\times\) at c56 and \(1.44\times\) at c64.

\paragraph{Pruning preserves acceptance while cutting verification cost.}
\dcut is only worthwhile if the positions it discards are ones the target would have rejected anyway, and the acceptance measurements confirm that they are.
Averaged over the sweep, mean acceptance length is \(2.50\) for \dflaretwo and \(2.46\) for \dcut, a reduction of roughly \(1.5\%\).
Even at concurrency 64, where verification is most contended, acceptance stays within \(2.8\%\) of the unpruned baseline, moving from \(2.50\) to \(2.43\), while throughput rises by \(15.7\%\).
This asymmetry is exactly what \Cref{eq:dcut-score} is built to expose: a deep position whose prefix-survival estimate is small contributes almost nothing to \(\hat{A}_i\), so dropping it forfeits negligible expected progress yet removes a real verification slot from the batch.
\dcut thus reduces the target model's per-step work while leaving the drafter's realized acceptance essentially intact, which is why the throughput gain is far larger than the acceptance it gives up.
\section{Modular Performance Analysis}
\label{sec:modular_analysis}

\subsection{Motivation and Measurement}

End-to-end throughput alone does not identify the source of a performance difference between speculative decoding methods. Engine TPS depends jointly on the number of tokens committed in each decode iteration and the time required to complete that iteration. A method that commits more tokens per iteration may still achieve lower throughput if draft generation or target-model verification is sufficiently expensive. Component-level profiling is therefore needed to connect proposer design, iteration cost, committed length, and engine-level TPS.

We profile steady-state decoding under fixed request concurrency using NVIDIA Nsight Systems. All experiments are conducted on a single node equipped with eight NVIDIA H20 GPUs, each with 96\,GB of memory. Each experiment is warmed up for 60 seconds and then profiled for 4 seconds. Requests are sampled from MATH-500 with a fixed random seed of 1234, temperature 0, and a maximum generation length of 128 tokens. All evaluated configurations use the same Hy3 target model and a tensor-parallel size of 8, distributing model execution across all eight GPUs.

The comparison includes Hy3-MTP3 with three draft tokens; Hy3-DFlash-L3-B5 with four draft tokens; Hy3-DFlash-L5-B8 and Hy3-DFlash-L7-B8 with seven draft tokens; and three configurations from the Hy3-\dflaretwo family, each with seven draft tokens. These configurations are the Hy3-\dflaretwo backbone, the backbone augmented with a Markov head, and the final Hy3-\dflaretwo configuration using hidden correction.

In the Hy3-DFlash naming convention, \texttt{L$x$} denotes a proposer with $x$ model layers, while \texttt{B$y$} denotes an inference block size of $y$. For example, Hy3-DFlash-L3-B5 uses a 3-layer proposer and an inference block size of 5, corresponding to four draft tokens per iteration. Hy3-DFlash-L5-B8 and Hy3-DFlash-L7-B8 use 5-layer and 7-layer proposers, respectively, with an inference block size of 8, corresponding to seven draft tokens per iteration.

The analysis uses the GPU-projected durations of two instrumented NVTX ranges. The \texttt{execute\_model} range covers target-model execution and associated verification work, whereas \texttt{sample\_tokens} covers proposer execution, draft generation, sampling, and related operations. Their sum is used as a device-side decode-step latency proxy rather than as an exact end-to-end iteration latency, since CPU scheduling, request management, queueing, and other host-side overheads are not included.

Median range durations are used throughout the component and break-even analyses to characterize typical steady-state execution cost while reducing sensitivity to occasional long-duration observations. The decode-step latency proxy is the sum of the separately reported medians of \texttt{execute\_model} and \texttt{sample\_tokens}; it should therefore be interpreted as a typical-step proxy rather than as the median of paired end-to-end iterations. The Avg.\ row in the break-even table is the arithmetic mean of the nine concurrency-specific thresholds computed from these median latency proxies.

\subsection{Relating Component Latency to Engine TPS}

At a fixed request concurrency, TPS is proportional to the number of committed tokens per iteration divided by the iteration latency. We assume that Hy3-MTP3 accepts two draft tokens on average and commits one additional target-model token, giving a baseline committed length of three. Using the median component-latency proxy, the minimum committed length required for method $m$ to match the predicted throughput of Hy3-MTP3 is

\begin{equation}
C^{\mathrm{BE}}_m(b)
=
3
\frac{
\widetilde{T}_{\mathrm{execute},m}(b)
+
\widetilde{T}_{\mathrm{sample},m}(b)
}{
\widetilde{T}_{\mathrm{execute},\text{Hy3-MTP3}}(b)
+
\widetilde{T}_{\mathrm{sample},\text{Hy3-MTP3}}(b)
},
\label{eq:break_even_committed_length}
\end{equation}

where $b$ denotes request concurrency, $\widetilde{T}$ denotes the median GPU-projected range duration, and $C^{\mathrm{BE}}_m(b)$ includes the target-model bonus token. Method $m$ is predicted to outperform Hy3-MTP3 when its measured mean committed length exceeds this threshold. If the threshold is greater than the maximum number of tokens that the method can commit in one iteration, the method cannot outperform Hy3-MTP3 at that operating point under the latency model.

Equation~\ref{eq:break_even_committed_length} connects typical component-level execution cost to the committed length required for a TPS improvement. A higher iteration cost raises the required committed length, whereas a lower iteration cost reduces it. This relationship remains a device-side prediction and must ultimately be validated using end-to-end engine measurements.

\subsection{Component Latency}

\begin{figure*}[t]
    \centering
    \begin{subfigure}[t]{0.49\textwidth}
        \centering
        \includegraphics[width=\linewidth]
        {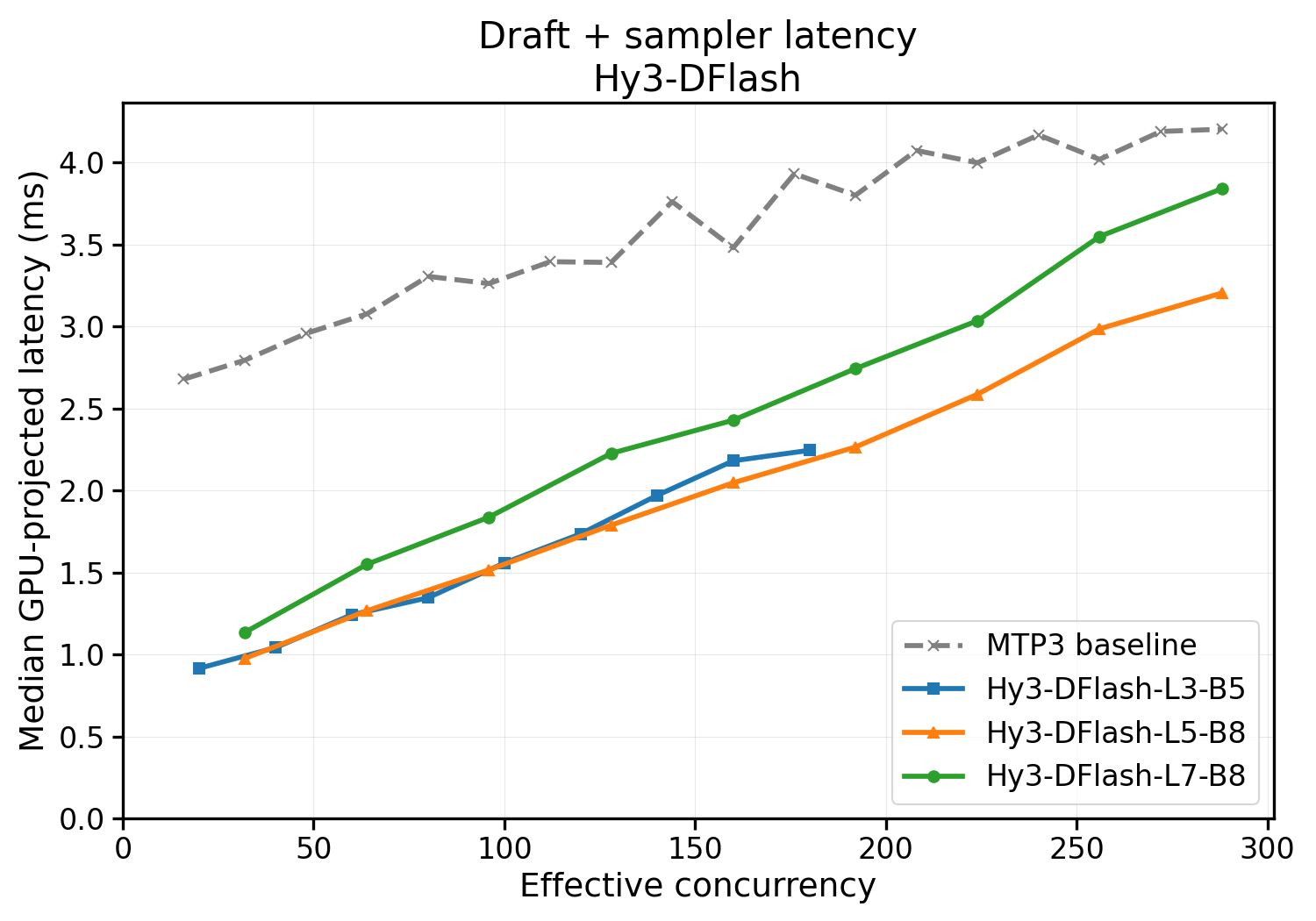}
        \caption{Hy3-DFlash: draft and sampler latency.}
    \end{subfigure}
    \hfill
    \begin{subfigure}[t]{0.49\textwidth}
        \centering
        \includegraphics[width=\linewidth]
        {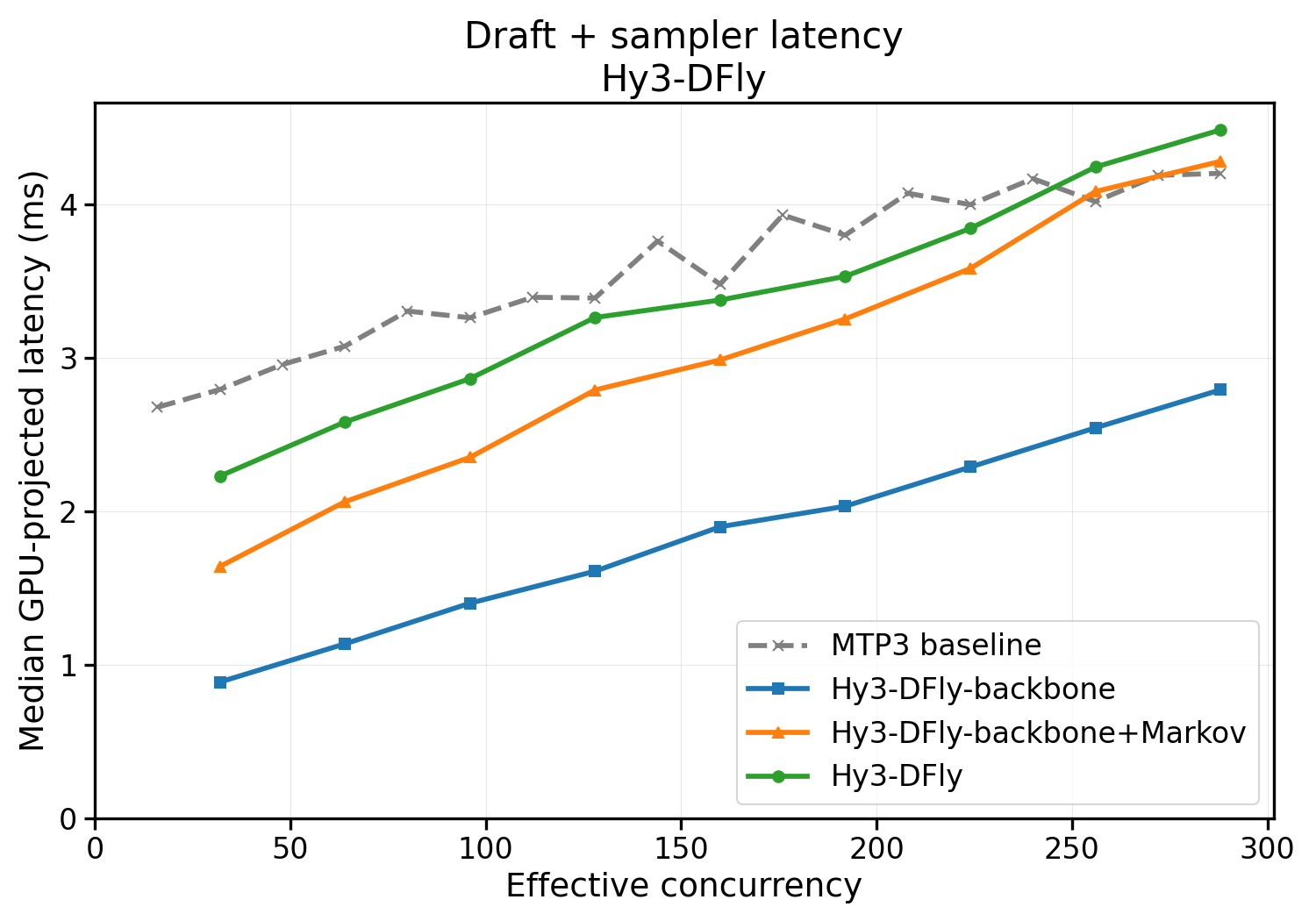}
        \caption{Hy3-\dflaretwo: draft and sampler latency.}
    \end{subfigure}

    \vspace{0.5em}

    \begin{subfigure}[t]{0.49\textwidth}
        \centering
        \includegraphics[width=\linewidth]
        {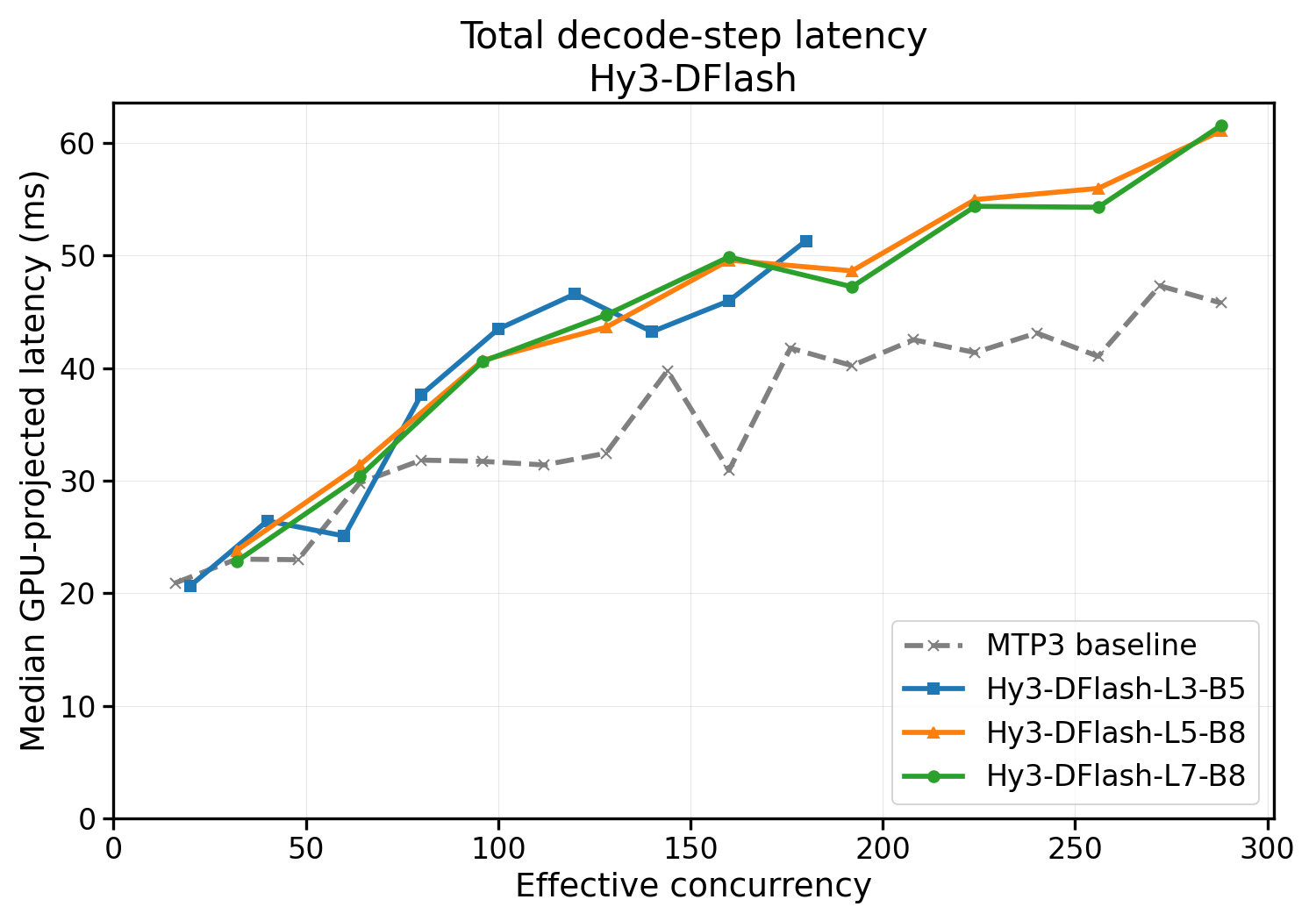}
        \caption{Hy3-DFlash: total decode-step latency proxy.}
    \end{subfigure}
    \hfill
    \begin{subfigure}[t]{0.49\textwidth}
        \centering
        \includegraphics[width=\linewidth]
        {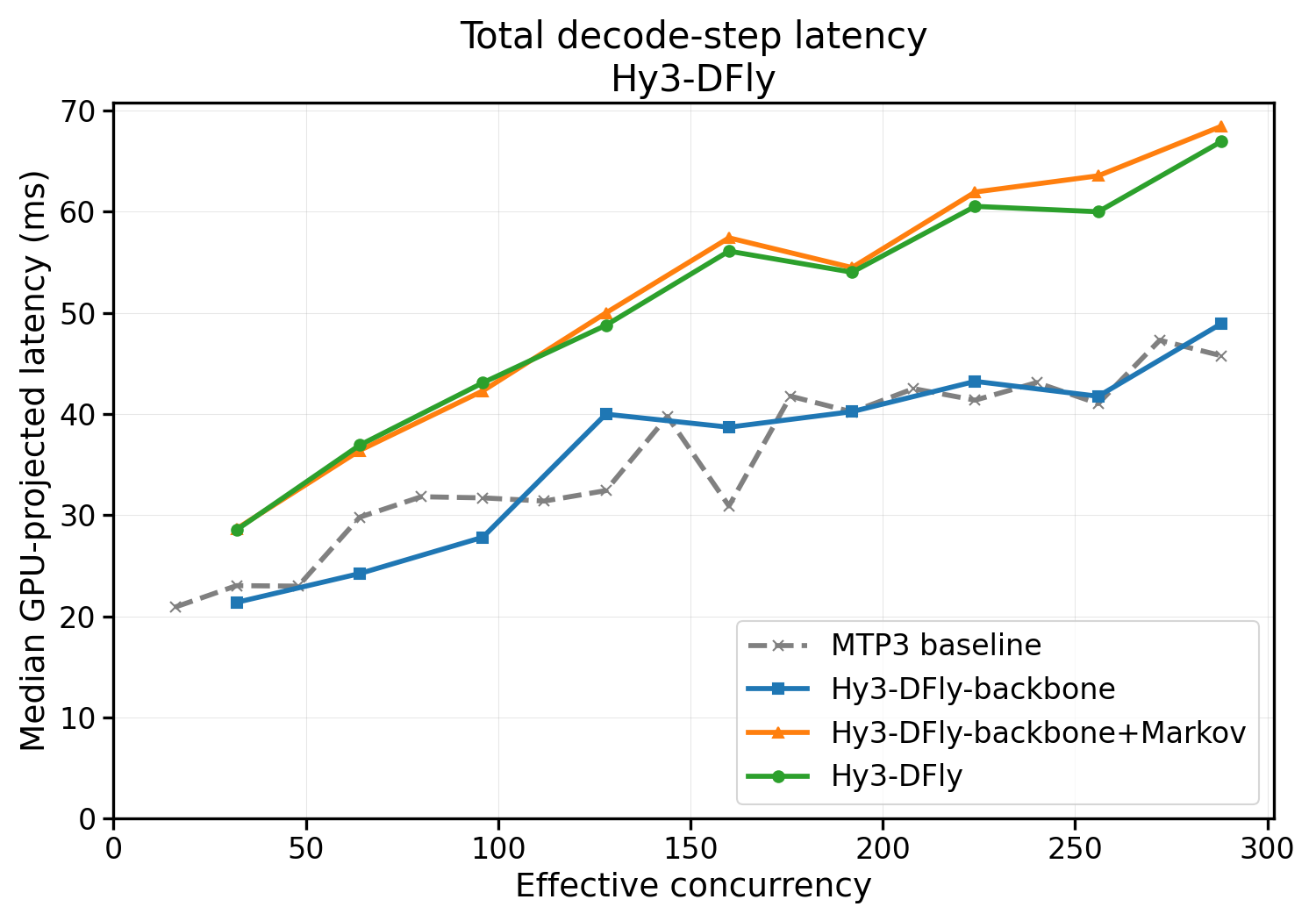}
        \caption{Hy3-\dflaretwo: total decode-step latency proxy.}
    \end{subfigure}

    \caption{Median GPU-projected component latency and total decode-step
    latency proxy. Effective concurrency is defined as $b(1+k)$, where $b$ is
    request concurrency and $k$ is the configured number of draft tokens.
    Because $k$ differs across configurations, points at the same effective
    concurrency do not necessarily correspond to the same request concurrency.
    Hy3-MTP3 is therefore shown as a timing reference rather than as a direct
    engine-speedup comparison.}
    \label{fig:hy_component_latency}
\end{figure*}

Figure~\ref{fig:hy_component_latency} shows that target-model execution is the dominant component of device-side latency. Across the updated Hy3 measurements, the median \texttt{execute\_model} duration ranges from approximately $19.77$ to $64.16$ ms, whereas the median \texttt{sample\_tokens} duration ranges from approximately $0.89$ to $4.49$ ms. Component analysis is therefore needed to determine whether the additional cost introduced by a larger proposer can be recovered through a longer committed sequence.

For Hy3-DFlash, expanding the design from Hy3-DFlash-L3-B5 to Hy3-DFlash-L5-B8 increases both the proposer depth and the inference block size, while the resulting increase in component latency remains manageable. Correspondingly, the average break-even committed length increases only from $3.80$ to $4.61$ tokens, an increase of $0.81$ tokens, while the maximum committed length increases from 5 to 8 tokens. The increase in the required committed length is therefore substantially smaller than the additional committed-length capacity introduced by the larger block. This provides useful headroom for improving the actual accepted length through architectural scaling.

The two configurations with an inference block size of 8, Hy3-DFlash-L5-B8 and Hy3-DFlash-L7-B8, have similar total latency proxies and average break-even committed lengths of $4.61$ and $4.56$ tokens, respectively. The difference of $0.05$ tokens indicates that increasing the proposer depth from 5 to 7 layers does not materially change the required committed length. Hy3-DFlash-L5-B8 nevertheless has lower draft-and-sampler latency at all nine evaluated request concurrencies. We therefore select Hy3-DFlash-L5-B8 as the primary DFlash configuration, retaining the larger inference block while limiting proposer-side overhead.

For the Hy3-\dflaretwo family, we focus on the comparison between the final Hy3-\dflaretwo configuration and Backbone + Markov. The final Hy3-\dflaretwo configuration has a lower total latency proxy at seven of the nine evaluated request concurrencies. Its average total latency proxy is approximately $50.54$ ms, compared with $51.46$ ms for Backbone + Markov. Thus, hidden correction can be incorporated without increasing the average total decode-step cost relative to the Markov configuration, motivating the selection of the final Hy3-\dflaretwo configuration for subsequent evaluation.

\subsection{Break-Even Committed Length}

Table~\ref{tab:hy_break_even_committed_length} reports the minimum committed length required to match the predicted TPS of Hy3-MTP3 at the same request concurrency. Each threshold is calculated from the median \texttt{execute\_model} and \texttt{sample\_tokens} durations. Unlike Figure~\ref{fig:hy_component_latency}, which uses effective concurrency for component-level visualization, the break-even calculation matches every method with Hy3-MTP3 at the same request concurrency.

\begin{table*}[t]
    \centering
    \small
    \setlength{\tabcolsep}{6pt}
    \renewcommand{\arraystretch}{1.2}
    \setlength{\aboverulesep}{0pt}
    \setlength{\belowrulesep}{0pt}
    \caption{\textbf{Break-even committed length across request concurrencies.}
    Each entry is calculated from the median GPU-projected durations of
    \texttt{execute\_model} and \texttt{sample\_tokens} and denotes the
    minimum committed length required to match the predicted throughput of
    Hy3-MTP3. The values include one target-model bonus token. Hy3-MTP3 is
    assumed to commit three tokens per iteration. Avg.\ is the arithmetic
    mean of the nine concurrency-specific thresholds. The configurations
    selected for subsequent evaluation are highlighted in bold in the second
    header row.}
    \label{tab:hy_break_even_committed_length}
    \resizebox{\textwidth}{!}{%
    \begin{tabular}{cccccccc}
        \toprule
        \multirow{2}{*}{\textbf{Request Concurrency}}
          & \multirow{2}{*}{\textbf{Hy3-MTP3}}
          & \multicolumn{3}{c}{\textbf{Hy3-DFlash}}
          & \multicolumn{3}{c}{\textbf{Hy3-\dflaretwo}} \\
        \cmidrule(lr){3-5}
        \cmidrule(lr){6-8}
          & & L3-B5
          & \textbf{L5-B8}
          & L7-B8
          & Backbone
          & Backbone + Markov
          & \textbf{Hy3-\dflaretwo} \\
        \midrule
         4 & 3.00 & 2.97 & 3.42 & 3.28 & 3.07 & 4.11 & 4.10 \\
         8 & 3.00 & 3.44 & 4.09 & 3.96 & 3.16 & 4.74 & 4.81 \\
        12 & 3.00 & 3.27 & 5.31 & 5.30 & 3.63 & 5.52 & 5.62 \\
        16 & 3.00 & 3.79 & 4.39 & 4.50 & 4.03 & 5.03 & 4.91 \\
        20 & 3.00 & 4.10 & 4.67 & 4.70 & 3.65 & 5.41 & 5.29 \\
        24 & 3.00 & 4.41 & 4.60 & 4.47 & 3.81 & 5.15 & 5.11 \\
        28 & 3.00 & 4.13 & 5.25 & 5.19 & 4.13 & 5.92 & 5.78 \\
        32 & 3.00 & 4.25 & 5.17 & 5.02 & 3.86 & 5.88 & 5.55 \\
        36 & 3.00 & 3.87 & 4.61 & 4.64 & 3.69 & 5.16 & 5.05 \\
        \midrule
        \textbf{Avg.}
          & \textbf{3.00}
          & \textbf{3.80}
          & \textbf{4.61}
          & \textbf{4.56}
          & \textbf{3.67}
          & \textbf{5.21}
          & \textbf{5.14} \\
        \bottomrule
    \end{tabular}%
    }
\end{table*}

Table~\ref{tab:hy_break_even_committed_length} converts the component-latency measurements into target committed lengths at different request concurrencies. The reported values include one target-model bonus token and can therefore be compared directly with the measured mean number of committed tokens per iteration. A method is expected to outperform Hy3-MTP3 under the device-side latency model only when its measured committed length exceeds the corresponding threshold.

For Hy3-DFlash, expanding from Hy3-DFlash-L3-B5 to Hy3-DFlash-L5-B8 increases the average break-even committed length from $3.80$ to $4.61$ tokens, while increasing the maximum committed length from 5 to 8 tokens. The structural expansion therefore increases the average committed-length requirement by only $0.81$ tokens while providing three additional positions of committed-length capacity. If the larger model improves the actual committed length by more than this additional requirement, the expansion is expected to produce a positive TPS benefit.

Hy3-DFlash-L5-B8 and Hy3-DFlash-L7-B8 have nearly identical average target committed lengths of $4.61$ and $4.56$ tokens. Hy3-DFlash-L5-B8 therefore needs to achieve a committed length close to that of Hy3-DFlash-L7-B8 to provide comparable TPS potential, while retaining lower proposer-side latency. This relationship motivates the selection of Hy3-DFlash-L5-B8 as the preferred trade-off between model cost and committed-length capacity.

For the Hy3-\dflaretwo family, Backbone + Markov and the final Hy3-\dflaretwo configuration require average committed lengths of $5.21$ and $5.14$ tokens, respectively. The final Hy3-\dflaretwo configuration therefore has a slightly lower target committed length, consistent with its lower total latency proxy at most evaluated request concurrencies. Its hidden-correction mechanism is expected to improve TPS when the measured committed length exceeds the corresponding concurrency-specific threshold.

The Avg.\ row provides a compact summary across request concurrencies, while operating-point-specific comparisons should use the threshold at the corresponding request concurrency. For the two selected configurations, Hy3-DFlash-L5-B8 and Hy3-\dflaretwo, the average target committed lengths are $4.61$ and $5.14$ tokens, respectively. These values define the committed-length targets implied by the measured component costs and can be compared with the observed committed lengths and end-to-end TPS in subsequent evaluation.

\section{AngelSpec Framework}
\label{sec:angelspec}
All drafters in \Cref{sec:mtp,sec:dflare} are trained with our
open-source training framework, built on top of
TorchSpec~\citep{torchspec2026}. We adopt TorchSpec for its disaggregated
foundation: inference engines run the target model and stream its hidden
states through a Mooncake-backed~\citep{qin2024mooncake} RDMA store directly to distributed
training workers, so data generation and optimization scale independently
and neither side stalls waiting on the other. The stack is torch-native,
and hidden states are captured inside vLLM worker processes through
public vLLM APIs---a speculative hidden-state extraction hook and a
custom KV connector---without forking the engine. TorchSpec, however,
centers on the EAGLE-3 and DFlash-family recipes; supporting the methods
of this report---the MTP drafter trained with TTT, the DFly family, and
long-context training on the Hy3 targets---requires extensions
throughout the pipeline. We implemented and compared a wide range of
drafter designs on this framework; the released version corresponds to
the best-performing configurations.

\subsection{MTP Drafter Support}
\label{sec:framework-mtp}

TTT (\Cref{sec:mtp-ttt}) requires an on-policy rollout inside the
training step: each prediction depth consumes the previous depth's own
prediction, a structure the standard teacher-forced training loop cannot
express. Our implementation unrolls the rollout in parallel over the
whole sequence. Each depth's forward pass appends one set of keys and
values to a depth-indexed cache, and the queries of every later depth
attend, in a single softmax, over the full causal prefix together with
the diagonal keys produced by earlier depths at the same position: the
prefix term runs through compiled FlexAttention, each diagonal term is a
per-position scalar dot product, and the terms are merged via logsumexp.
The causal-prefix-plus-diagonal attention structure used when drafting is
thus reproduced implicitly, without materializing TTT-specific masks or
full attention score matrices, and the memory of the multi-step rollout
stays close to that of a single causal pass.

\paragraph{Long-context training.} At long context, MTP training is
dominated by the memory of full-vocabulary logits and the activations of
the multi-step rollout, which together exceed single-device capacity well
before 128k tokens. The framework therefore trains MTP with Ulysses
sequence parallelism: the sequence is sharded across sequence-parallel
ranks, and each local shard carries a $D$-token halo so that the
depth-shifted supervision of every prediction depth remains rank-local
rather than requiring communication at shard boundaries; loss statistics
are then reduced across the sequence-parallel group. We validate this
path at context lengths up to 128k tokens.
\begin{figure}
    \centering
    \includegraphics[width=1\linewidth]{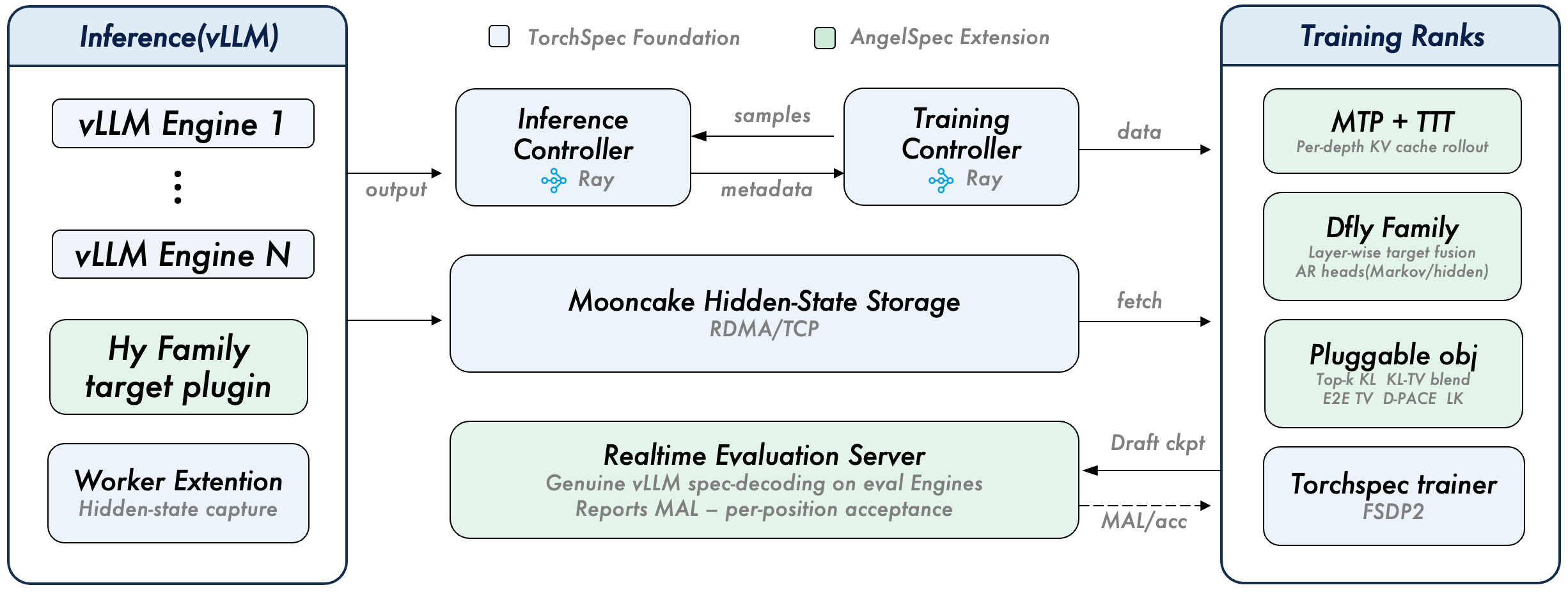}
    \caption{AngelSpec Overview.}
    \label{fig:placeholder}
\end{figure}
\subsection{DFlash-Family Training}
\label{sec:framework-dflash}

The framework trains the full family of \Cref{sec:dflare}. For target
conditioning, each draft layer holds learnable fusion weights over the
extracted target layers and forms its own softmax-weighted combination of
their features, which is added to the shared projected context to give
every draft layer a depth-differentiated conditioning representation. The
fusion weights are initialized with a distance prior, so that each draft
layer initially focuses on target layers at a matching depth and learns
to deviate from this assignment during training. The two autoregressive
heads---the low-rank Markov transition head and the hidden-correction
head---attach through the same component interface, with their structure
described in \Cref{sec:dflare}. Across the family, the vocabulary
embedding and language-model head are loaded from the target checkpoint
and kept frozen, so training updates only the draft backbone and the
added components.

\subsection{Features and Extensibility}
\label{sec:framework-features}

\paragraph{Document-aware sequence packing.} Draft training samples are
short blocks, so fixed-length batches waste most of their capacity on
padding. Similar to Megatron-style pretraining, the framework packs
multiple documents into fixed-length rows with a greedy first-fit
assignment, but keeps packed documents strictly isolated through three
mechanisms. First, an attention document gate restricts every query to
keys from the same document, so packed neighbors are mutually invisible.
Second, specific to the MTP path, a depth-shift document gate
progressively masks positions near the end of each document whose
depth-shifted labels would fall in the next document, so that no depth is
ever supervised across a document boundary; the DFlash path performs
single-step block prediction without such shifts and needs no analogue.
Third, positions are encoded document-locally, so each packed document
sees the same positional layout as if it were alone in the sequence.
Cross-document isolation is covered by unit tests that verify zero
attention leakage. Packing is supported on both the DFlash and MTP
training paths.

\paragraph{Evaluation server.} Proxy training metrics such as loss and
top-$k$ accuracy do not always track the acceptance behavior that
determines real speedup, and measuring the latter traditionally requires
exporting a checkpoint and standing up a dedicated serving environment.
The framework includes an evaluation server that periodically runs
genuine speculative decoding against the latest checkpoint on dedicated
evaluation GPUs and reports the resulting deployment-side metrics---mean
accepted length and per-position acceptance rates---as measured by the
serving engine itself. Two deployment modes are provided: a lightweight
mode that launches a fresh vLLM instance per evaluation, suitable for
single-node runs, and a persistent mode in which the target model stays
resident and only the draft weights are reloaded for each round, keeping
the recurring cost of evaluation small in multi-node training.

\paragraph{Extensibility across the training stack.} Exploring the
design space of the preceding sections requires that new components be
cheap to add, and the framework exposes pluggable interfaces at three
levels, each exercised in this work. \emph{Targets:} hidden-state
extraction for a new target model is injected into vLLM at runtime
through a plugin entry point, with no fork or source patch of the
engine---the Hy3 series is supported this way across both the
HuggingFace and vLLM backends. \emph{Objectives:} the loss family of
\Cref{sec:mtp-loss}---hard-label cross-entropy, top-$k$ KL distillation,
the adaptive KL--TV blend, and the end-to-end accepted-length
variant---together with the D-PACE-weighted hybrid LK objective of
\Cref{sec:dflare} compose over a shared base and are selected through
configuration rather than trainer forks. \emph{Optimizers:} Muon is
available as a drop-in alternative. We additionally hardened the
disaggregated pipeline for real clusters (multi-node, multi-NIC, and
proxied hosts) and support cross-node tensor-parallel inference for
large targets. These extensions turn the base trainer into a general
platform for drafter research: the methods of this report were developed
on it, and we release it in the same form to support subsequent
exploration by the community.
\section{Related Work}
\label{sec:related_work}

Speculative decoding accelerates autoregressive generation through a lossless \emph{draft-then-verify} paradigm~\citep{sps1, sps2}, and a long line of work improves the drafter, some methods improve draft quality by reusing information from the target model, including intermediate hidden states, token embeddings, and additional multi-token prediction features~\citep{eagle,eagle2,eagle3,glide,mtp}, others from token trees~\citep{specinfer, sequoia, opt-tree, logitspec}; speculation has also been extended beyond the token level, e.g., drafting entire reasoning steps for the target model to verify~\citep{speccot}.
Target-aware drafters of this kind reduce the mismatch between the two models, but their proposals remain autoregressive: generating $\gamma$ draft tokens requires $\gamma$ dependent decoding steps, so drafting latency grows with proposal length. To remove the sequential bottleneck of drafting itself, a line of research proposes parallel or blockwise generation~\citep{medusa,p-eagle,pard,dart,dflash,zhang2026dflare}. 
However, their positions cannot condition on realized tokens within the same block, leading to suffix acceptance decay. Some efforts address this limitation by adding lightweight causal modules to parallel representations~\citep{huang2026domino,dspark,rheinboldt2026treeflashparallelarapproximationfaster}. Beyond drafter architecture, draft quality also depends on how training aligns the proposal distribution with the target model, extensive works focus on better draft--target alignment through on-policy data, tailored divergences, and online knowledge distillation~\citep{zhou2023distillspec,liu2023online,lk-losses,wu2026d}.

At serving time, a fixed proposal or verification depth is suboptimal because acceptance varies across domains, requests, and decoding steps, while the cost of verification changes with batch size and online load.
Most prior adaptive methods change the draft length or tree online to avoid wasting verification on low-confidence candidates~\citep{specdec++,disco,talon,specbranch,dyspec,pearl,zhang2026learning,shen2026draft}.
These methods use confidence, recent acceptance, or learned policies to adjust speculation, but request-local decisions do not fully capture competition for shared target-model capacity under batched serving.
DSpark~\citep{dspark} and D-cut~\citep{liu2026d} instead treats verification tokens across concurrent requests as a global resource and selects retained prefixes by combining their expected acceptance utility with hardware-profiled runtime cost.

\section{Conclusion}
\label{sec:conclusion}

We presented \method, a unified framework for training and evaluating speculative-decoding methods under heterogeneous real-world workloads.
Rather than relying on one universal drafter, \method specializes both model structure and training data: autoregressive MTP drafting targets high-entropy conversation, while block-diffusion drafting is strengthened with code- and mathematics-focused data for longer predictable continuations.
For the latter, we proposed \dflaretwo, which combines a hybrid target-conditioning backbone with a predecessor-conditioned autoregressive head to improve target-feature utilization and intra-block dependency modeling while retaining a parallel backbone.
At inference time, the system treats verification as a shared batch-level resource and adapts retained depth by combining expected prefix utility with hardware-profiled runtime cost. On Hy3-A20B, \dflaretwo achieves the highest average throughput at every tested concurrency from 4 to 64, delivering a $1.98$--$2.40\times$ speedup over autoregressive decoding and $10.5$--$11.8\%$ higher throughput than DFlash. 
We also release the \method framework with unified support for MTP and block-parallel speculative decoding, providing a common foundation for training, evaluating, and extending future drafter designs. On Hy3-A20B, \dflaretwo increases the average accepted length by approximately $30\%$, enabling the highest average throughput at every tested concurrency from 4 to 64, delivering a $1.98$--$2.40\times$ speedup over autoregressive decoding and $10.5$--$11.8\%$ higher throughput than DFlash. 
We also release the \method framework with unified support for MTP and block-parallel speculative decoding, providing a common foundation for training, evaluating, and extending future drafter designs.

\newpage

\bibliographystyle{citation}
\bibliography{citation}

\newpage
\appendix

\end{document}